%% file: main.tex
\PassOptionsToPackage{table}{xcolor}
\documentclass[11pt]{article}
\usepackage{caption}
\usepackage{makecell}
\usepackage{array}
\usepackage{enumitem}
\usepackage{textcomp}
\usepackage{CJKutf8}

\usepackage{amsmath}
\usepackage{amssymb}
\usepackage[most]{tcolorbox}
\tcbuselibrary{breakable, skins}
\usepackage{listings}
\usepackage{alltt}

\usepackage{subcaption}

\usepackage{booktabs}
\usepackage{graphicx}
\usepackage{multirow}
\usepackage{xcolor}
\usepackage{colortbl}
\definecolor{lightgrayrow}{RGB}{235,235,235}
\usepackage{arydshln}

\usepackage[final]{acl}

\usepackage{times}
\usepackage{latexsym}

\usepackage[T1]{fontenc}

\usepackage[utf8]{inputenc}

\usepackage{microtype}

\usepackage{inconsolata}

\usepackage{graphicx}

\title{LiveSim: Simulating Environment-Shaped Users in Multi-Agent Live-Stream Ecosystems}

\author{
 \textbf{Jiaqi Xu\textsuperscript{1}}\thanks{Work was conducted during the internship at ByteDance.}\thanks{Jiaqi Xu and Yiran Qiao contributed equally to this work.},
 \textbf{Yiran Qiao \textsuperscript{1}}\footnotemark[1]\footnotemark[2],
 \textbf{Jing Chen\textsuperscript{2}},
 \textbf{Qiwei Zhong\textsuperscript{2}},
 \textbf{Xiang Ao\textsuperscript{1}}\thanks{Xiang Ao is the corresponding author.},
 \textbf{Xueqi Cheng\textsuperscript{1}}
\\
 \textsuperscript{1} University of Chinese Academy of Sciences, Beijing 100049, China \\
 \textsuperscript{2} ByteDance China\\
    \{jiaqxv, yrqiao\}@gmail.com \\
    \{yilan.chan, huafeng.hf\}@bytedance.com \\
    \{aoxiang, chengxueqi\}@ict.ac.cn
}

\begin{document}
\maketitle
\begin{abstract}
\input{body/abstract}
\end{abstract}

\input{body/intro_new}
\input{body/methodology_new_new}
\input{body/experiments_new}
\input{body/relatedwork_new}
\input{body/conclusion}
\input{body/acknowledgements}

\section{Limitations}
Our experiments are conducted on live-streaming risk trajectories from a specific platform setting. Although the framework is designed to be general, its effectiveness on other domains such as e-commerce chat, short-video comments, or open social networks remains to be further validated. The behavioral hypothesis used by LiveSim is inferred from sparse interaction logs and should not be interpreted as the user's true intent or psychological state. It is an editable simulation hypothesis rather than a ground-truth user profile.

\section*{Ethical Considerations}
LiveSim is built on de-identified live-streaming interaction logs collected under the source platform's terms of service. We do not release the raw logs, and all identifiers, usernames, and free-text fields are removed or hashed before any prompt is constructed. As LiveSim models user susceptibility and risk migration, it should be used only for safety analysis, moderation evaluation, and protective intervention design. It should not be used to optimize persuasive manipulation or increase user engagement with risky content. Our use of commercial LLM backbones (Doubao, Qwen, DeepSeek and GPT) for both simulation and judging complies with the providers' acceptable-use policies; prompts and outputs that involve fraudulent-persuasion content are confined to offline evaluation runs and are not deployed in any user-facing setting.

\nocite{mou2025agentsense,shengbinyue2025multi,liu2025mosaic,yao2025dischargesim,zhao2023competeai,wang2024simulating,mou2024individual,mou2024unveiling,pryzant2023automatic,cui2024introducing,qiao2026live,qiao2026deja,qiao2026outsmarting,lian2025panoramic}

\bibliography{custom}

\appendix
\input{body/appendix}

\end{document}

%% file: body/abstract.tex
User behavior simulation with large language models~(LLMs) is increasingly used to support multi-agent ecosystem simulation. Existing simulators typically rely on static user profiles inferred from historical observations, which become inadequate in socially intensive environments such as live streaming where interaction dynamics continuously reshape user behavior.
We propose \textbf{LiveSim}, an LLM-based framework for live-stream ecosystem simulation. It represents users as editable behavioral hypotheses and progressively refines them through trajectory-grounded interactions, where discrepancies between simulated and observed trajectories reveal missing environmental shaping effects. These signals are further extracted as transferable environment-behavior patterns and accumulated in a collective behavioral memory to improve user-level behavioral fidelity and support ecosystem-level simulation.
Experiments on real-world live-stream risk-control data validate the effectiveness of LiveSim in improving user-level behavioral fidelity and enabling ecosystem-level analysis of risk evolution and platform intervention effects.

%% file: body/intro_new.tex
\section{Introduction}

\begin{figure}[t]
    \centering
    \includegraphics[width=1\linewidth]{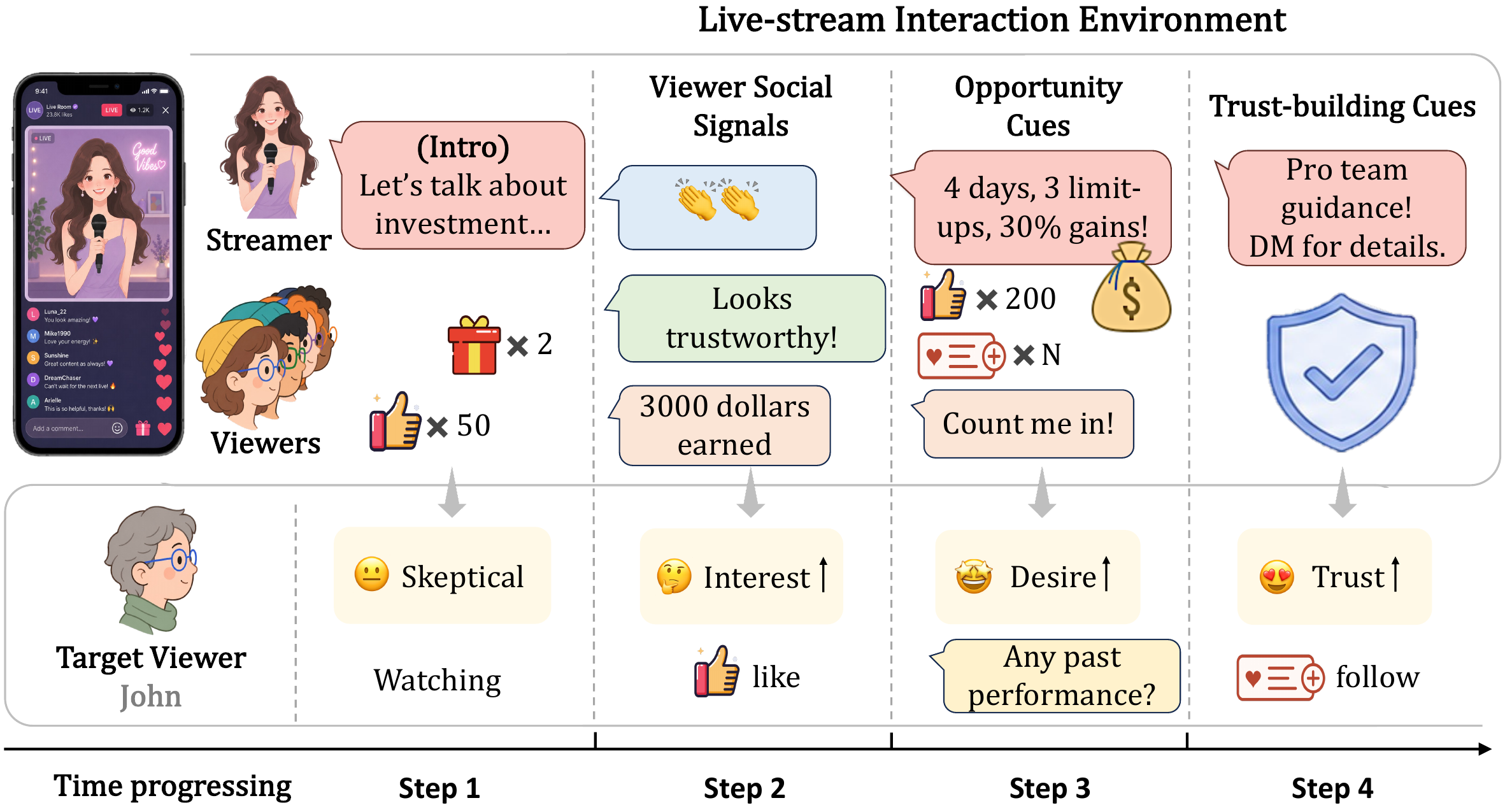}
    \caption{In our live-stream setting, users correspond to viewers whose behaviors are shaped by interaction environments, including streamer persuasion, viewer social signals, and evolving live-stream session dynamics.}
    \label{fig:intro}
\end{figure}

Large language models (LLMs) have enabled increasingly sophisticated simulations of human behavior, from individual decision-making~\cite{duan2026lifesim,bougie2025simuser} to multi-agent social ecosystems~\cite{park2023generative,yang2024oasis,mou2024unveiling,huang2026policysim}. Existing user simulators~\cite{bougie2025simuser,wang2025user,zhu2025llm,wang2025know,dai2025profile} typically focus on modeling user behaviors under interaction environments using predefined user descriptions. However, they place less emphasis on explicitly modeling environment-driven behavioral dynamics. While effective in relatively stable settings such as virtual societies and social networks~\cite{park2023generative,yang2024oasis,mou2024unveiling,tang2025gensim}, this paradigm fundamentally breaks down in socially intensive environments. For example, in live streaming, streamer persuasion, real-time viewer feedback, and evolving session dynamics continuously reshape user behavioral tendencies within a single session~(c.f.~Figure~\ref{fig:intro}).

This creates a core challenge: even for users with sufficient interaction history, historical logs provide only sparse snapshots of past actions, capturing what users eventually did but not how their behaviors formed under interaction. As a result, user characterizations inferred from history are inherently underspecified, i.e., insufficient to determine how users respond to unseen interaction conditions. We therefore reconceptualize user characteristics as \textbf{editable behavioral hypotheses}: provisional assumptions inferred from available evidence, subject to revision as interactions unfold.

Under this view, simulation becomes a process of progressively shaping behavioral hypotheses rather than repeatedly sampling actions from fixed user profiles. Discrepancies between simulated and observed trajectories are not merely errors to minimize. They are diagnostic signals revealing where current hypotheses fail to capture how environments reshape users. While adapting hypotheses from isolated trajectory failures risks instability, as individual trajectories are tightly coupled to specific session contexts. We observe that recurring interaction conditions tend to influence users in consistent ways, motivating the extraction of transferable \textbf{environment–behavior shaping patterns} that generalize across users and sessions.

To address these challenges, we propose \textbf{LiveSim}, an LLM-based framework for hierarchical simulation in live-stream ecosystems. LiveSim requires only these lightweight interaction logs and coarse user or session metadata. LiveSim first operates at the user level by initializing users from behavioral hypotheses inferred from sparse histories and progressively shaping the hypotheses through trajectory-grounded interactions. When behavioral hypotheses fail to explain observed trajectories, LiveSim employs LLM-based reflection to extract environment–behavior patterns describing how specific interaction conditions reshape user states and behavioral tendencies. These shaping patterns are accumulated into a collective behavioral memory and reused across users as transferable behavioral priors. Built upon improved user-level behavioral fidelity, LiveSim further deploys shaped user agents into closed-loop multi-agent ecosystems for studying interaction dynamics and platform interventions.
Experiments on real-world live-stream risk-control data from a major platform demonstrate that LiveSim substantially improves behavioral fidelity at the user level and improves ecosystem-level simulation quality for studying platform interventions, validating both the hypothesis-shaping paradigm and its utility for ecosystem-level analysis.

From a broader perspective within live-stream ecosystems, achieving individual-level behavioral fidelity is not an end in itself, but rather the cornerstone for building a trustworthy, reliable, and controllable ecosystem-level simulation. By bridging the gap between micro-level psychological shifts and macro-level system safety, LiveSim provides a high-fidelity sandbox for proactively prototyping platform interventions and testing risk-control strategies. By precisely capturing how dynamic interaction environments reshape user responses, we can robustly evaluate the emergent derivative risks within live sessions, as well as the true efficacy of protective countermeasures.

%% file: body/methodology_new_new.tex
\begin{figure*}[t]
    \centering
    \includegraphics[width=\textwidth]{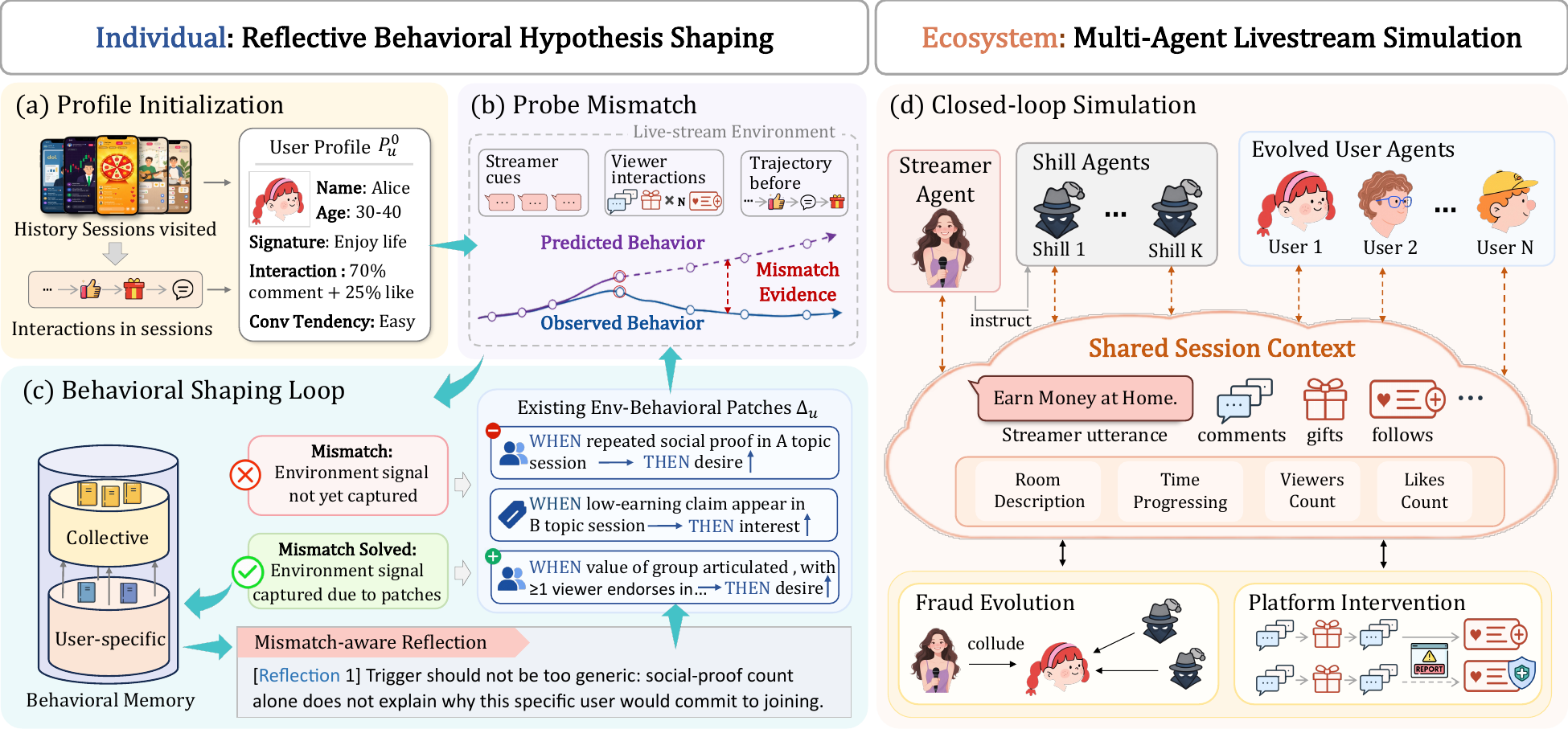}
    \caption{\textbf{Overview of LiveSim.}
    \textbf{(Left) Individual-level RBHS}: an initial hypothesis $P_u^0$ is iteratively shaped by mismatch evidence into environment--behavior patches, with reflections accumulated as a \emph{Behavioral Memory} for transfer.
    \textbf{(Right) Ecosystem-level Multi-Agent Simulation}: evolved user agents interact with streamer and shill agents in a closed-loop session for studying fraud evolution and platform intervention.}
    \label{fig:main}
\end{figure*}

\section{LiveSim Framework}
\label{sec:method}

\subsection{Framework Overview}

\textbf{LiveSim} realizes the editable-hypothesis view at two hierarchical levels, as shown in Figure~\ref{fig:main}.

\textbf{Individual level: Reflective Behavioral Hypothesis Shaping
(\S\ref{sec:rbhs}).}
For each user, LiveSim starts from an initial behavioral hypothesis
inferred from sparse history, then exposes it to live-stream session signals to
surface where the hypothesis fails to explain observed behavior. A
reflective shaping loop turns these mismatches into environment-conditioned
behavioral updates that evolve the hypothesis without rewriting the user's
underlying identity. Effective shaping patterns are further accumulated
into a collective memory and reused as transferable priors for subsequent
users.

\textbf{Ecosystem level: Closed-loop multi-agent simulation
(\S\ref{sec:ecosystem}).}
The evolved user agents are then deployed into a closed-loop livestream
session together with a streamer agent and shill agents, all interacting
through a shared session context. Individual-level fidelity propagates into
session-level dynamics, enabling controlled study of fraud evolution under
varying shill compositions as well as the effectiveness of platform
interventions.

\subsection{Reflective Behavioral Hypothesis Shaping}
\label{sec:rbhs}

Rather than treating simulation mismatches as failures, we view them as supervision. Building on this view, we design \textbf{Reflective Behavioral Hypothesis Shaping (RBHS)}, a three-stage process that converts these mismatches into effective environment--behavior patches to improve individual behavioral fidelity in simulation.

\subsubsection{Initial Behavioral Hypothesis}
\label{sec:hypothesis}
For each user $u$, an initial behavioral hypothesis $P_u^0$ is inferred from sparse historical evidence, including lightweight user metadata, previously visited sessions, past actions, and comment content. We organize the inferred attributes into three groups: identity cues extracted from the user's nickname and signature; interest preferences derived from recurring session categories and interaction habits; and risk-relevant tendencies such as skepticism toward exaggerated claims or willingness to follow a streamer. Full inference details are deferred to Appendix~\ref{sec:appendix_RBHS}.

\subsubsection{Probe Mismatch}
\label{sec:probe}
While $P_u^0$ provides an initial prior over the user, it describes \emph{who $u$ tends to be} but not how $u$'s behavior is shaped by environmental stimuli, e.g., which streamer cues build trust, or which signals invite skepticism. A natural way to expose such gaps would be to simulate $u$ over the full historical trajectory under the current hypothesis and compare the simulated actions with the real ones. This is impractical because live-stream session simulation is strictly serial. Specifically, each step depends on the updated state and memory from previous steps, so repeated full-trajectory simulation scales prohibitively with trajectory length.

We therefore use \emph{behavioral probes}: rich-context one-step prediction windows extracted from the historical trajectory. For a target step, the probe gives the LLM the current session stimulus together with sampled history from the same trajectory, including streamer utterances, salient session signals, audience comments, and the user's earlier actions. The probe then asks for a single next action under $P_u$.
The sampled context serves as a proxy for the state trajectory that step-by-step simulation would produce, letting one-step probes approximate full rollouts without simulating every intermediate step.
Many probe points can be batched into one LLM call to yield dense (predicted, observed) pairs.

A \emph{mismatch} is recorded when the probed action diverges from the observed action at the same step, and is summarized as a short \emph{mismatch signature}, e.g., predicting a like where $u$ commented, or missing a follow after sustained social proof, etc. Downstream shaping consumes the signature rather than the raw error. This localizes the environmental condition that the current $P_u$ fails to explain, turning each disagreement into a targeted cue for environment--behavior patching.

\subsubsection{Behavioral Shaping Loop}
\label{sec:Behavioral Shaping Loop.}
 After mismatch signatures are generated, a \emph{Behavioral Hypothesis Update Module} reads discrepancies as a diagnostic question: \emph{what is missing from $P_u$ that would have produced the observed action under this context?} Answers are viewed as environment-behavior patches $\Delta_u$ as follows.
\paragraph{Environment-behavior patch.} Each patch captures a single environment-triggered dynamic: A specific session condition should have shifted $u$'s latent state, which in turn would have reshaped action propensity. It is written in a \textsc{When}--\textsc{Then} schema that pairs the localized trigger with the corresponding state shift. This form rests on a simple observation: surface actions are noisy realizations of a more stable internal state, so patching how the environment shifts state is more transferable than patching individual actions.

All generated patches are concise natural-language statements that augment $P_u^0$, and are honored as priors in subsequent probes. Formally, RBHS evolves an initial hypothesis $P_u^0$ into
$\tilde{P}_u = P_u^0 \oplus \Delta_u$, where $\Delta_u$ is an additive
patch set accumulated from mismatch-driven shaping.

\paragraph{Reflection-driven shaping loop.} A single round of patching is rarely reliable: patches may misfire, over-strengthen a state dimension, or fail to improve alignment. RBHS therefore re-probes $u$ under the patched hypothesis $P_u^0 \oplus \Delta_u$ and asks the agent, at each probe step, to attribute its decision to the specific patches it relied on. Each probe yields a binary HIT score, and we summarize the change before and after patching as $\Delta\mathrm{HIT} \in \{-1, 0, +1\}$. For every patch we then count its associated probes: $n_{\mathrm{improved}}$, $n_{\mathrm{degraded}}$, and $n_{\mathrm{still\text{-}failing}}$ (probes whose HIT stays at $0$). A patch is labelled \emph{effective} when $n_{\mathrm{improved}} > 0$ and $n_{\mathrm{degraded}} = 0$, \emph{regressive} when $n_{\mathrm{degraded}} > 0$, and \emph{neutral} otherwise; neutral patches with $n_{\mathrm{still\text{-}failing}} > 0$ are also routed to bad evidence, since they fail to recover the user's behavior under the triggering context. Effective, regressive, and still-failing-bearing neutral patches, together with their probe contexts and predicted-versus-observed actions, are passed in a single LLM call that summarizes a compact set of \emph{reflections}: good reflections describe the user behavior pattern that effective patches captured, while bad reflections describe the pattern that should be avoided. These reflections guide the next patch generation round, where every newly written patch carries a marker \texttt{new} or \texttt{replace} with related patch ids, so that new patches are added as additional active rules or used to retire ineffective ones; non-replaced patches stay alive by default. The updated $P_u^0 \oplus \Delta_u$ is then re-probed. For each user the loop runs on a fixed schedule of three rounds, balancing cost and the depth needed to converge on concise patches.

\paragraph{Collective Behavioral Memory.} The reflection loop above presupposes prior rounds; a new user's first round has none. We bridge this cold start with a \emph{Collective Behavioral Memory} $\mathcal{M}$ of global reflections that capture recurring shaping trends rather than user-specific feedback, accumulated by the same Good/Bad summarization as user-level reflections but with evidence pooled across users. In each global round, a learner subset runs patch generation under the current $\mathcal{M}$ (empty in round $1$), and the resulting patches are scored by the same effective / regressive / neutral verdict. To summarize the most reusable pattern, we then pass the top $16$ effective patches by $n_{\mathrm{improved}}$, the top $16$ regressive or still-failing-bearing patches by $n_{\mathrm{degraded}} + n_{\mathrm{still\text{-}failing}}$, their probe contexts, and the current $\mathcal{M}$ to a single LLM call that emits the next batch of reflections. A new reflection identical to an existing one refreshes that entry as a strengthening signal; otherwise it is appended, and once $\mathcal{M}$ exceeds its budget of $6$, the oldest never-strengthened entries are retired first. Pooling runs on a learner subset of size $0.2 \cdot |\mathcal{U}|$ for $4$ rounds. At application time, a lightweight relevance gate makes one LLM call to admit the entries fitting the target user's $P_u^0$ and probe evidence, which then seeds first-round patch generation toward effective patterns and away from known failure modes.

\subsection{Closed-loop Multi-Agent Simulation}
\label{sec:ecosystem}
RBHS produces, for each user $u$, a shaped hypothesis $\tilde{P}_u = P_u^0 \oplus \Delta_u$ that captures both who $u$ tends to be and how $u$'s behavior is shaped by environmental stimuli. We now place these shaped agents back into a live-stream session and let them, together with a streamer agent, generate the session interaction step by step.

\paragraph{Session Composition.}
A simulated session replays the membership of a real historical session: the audience consists of the exact set of users who attended that session, each instantiated as an evolved agent carrying its own $\tilde{P}_u$. The streamer is driven by a hypothesis $P_{\text{str}}$ summarized from real streamer behavior on the platform and held fixed throughout simulation; unlike audience agents, the streamer is not subject to RBHS, since our focus is on how environments shape \emph{viewers}, and a stable streamer keeps the session-side condition controllable for downstream analysis.

\paragraph{Per-Step Dynamics.}
Each audience agent chooses from six actions: \textsc{like}, \textsc{comment}, \textsc{gift}, \textsc{follow}, \textsc{private-message}, and \textsc{join-group}; \textsc{comment} additionally carries free-form text generated by the LLM. At step $t$, let $c_t$ denote the session context aggregating recent streamer utterances, salient session signals, and audience activity produced up to step $t$. For each audience agent $u$, we maintain a cognitive state $s_t^u \in \{\text{interest}, \text{trust}, \text{desire}, \text{fatigue}\}$ and an interaction memory $\mathcal{H}_t^u$ recording $u$'s own past actions together with the session signals $u$ has observed. Conditioned on the shaped hypothesis, the current state, and memory, the agent jointly produces a next action and an updated state in a single LLM call:
\[
\begin{aligned}
(a_t^u,\; s_{t+1}^u)
&\sim \pi_\Theta\!\left(c_t,\; \tilde{P}_u,\; s_t^u,\; \mathcal{H}_t^u\right), \\
\mathcal{H}_{t+1}^u
&\leftarrow \mathrm{Update}(\mathcal{H}_t^u, c_t, a_t^u).
\end{aligned}
\]
Coupling action and state in one decoding step lets the hypothesis act on the latent state directly, rather than first committing to an action and then back-rationalizing the state. The streamer agent follows the same template under $P_{\text{str}}$, but without an evolving cognitive state: its next utterance is produced from $\pi_\Theta(c_t, P_{\text{str}}, \mathcal{H}_t^{\text{str}})$.

\paragraph{Closing the Loop.}
Audience actions and streamer utterances at step $t$ are committed to the session context $c_{t+1}$, which the next step exposes back to every agent. Through this shared context, one viewer's comment can shift another viewer's trust, and a streamer's persuasion attempt can propagate through audience reactions before circling back as new pressure on the streamer. Session dynamics emerge from agent--agent and agent--streamer coupling rather than being prescribed.

%% file: body/experiments_new.tex
\section{Experiments}

\newcommand{\calib}{\hspace{1em}\textit{+ RBHS}  (\(\Delta\))}
\newcommand{\na}{\textemdash}

\begin{table*}[t]
\centering
\small
\setlength{\tabcolsep}{5pt}
\renewcommand{\arraystretch}{1.12}

\caption{Overall performance of LiveSim across different backbones, evaluated under the trajectory-grounded protocol. Best in bold, second-best underlined.}
\label{tab:may_result}

\begin{tabular*}{\textwidth}{@{\extracolsep{\fill}} l c c c @{\hspace{10pt}} c c @{\hspace{10pt}} c c c @{}}
\toprule
\multicolumn{1}{c}{\multirow{2}{*}[-0.5ex]{\makecell[c]{\strut Models}}}
& \multicolumn{3}{c}{Micro}
& \multicolumn{2}{c}{Macro}
& \multicolumn{3}{c}{LLM-Judge} \\

\cmidrule(lr){2-4}\cmidrule(lr){5-6}\cmidrule(lr){7-9}
& HIT  & A-JSD $\downarrow$ &  Tr-JSD $\downarrow$
& Conv-Acc  & Conv-F1
& Align & Consist & Plaus\\
\midrule

Random
& 15.19 & 0.3390 & 0.5667
& 51.02 & 34.50
& 24.52 & 30.16 & 37.85 \\
\midrule
Qwen2.5-7B
& \underline{54.39} & 0.1551 & 0.3243
& 72.88 & 50.71
& 48.44 & 57.38 & 70.95 \\
\noalign{\vskip 2pt}
Qwen2.5-32B
& 54.28 & 0.0962 & 0.2089
& 78.37 &  45.76
& 52.23 & 62.26 & 73.82 \\
\noalign{\vskip 2pt}
GPT-4o-mini
& \textbf{57.03} & 0.1306 & 0.2525
& 75.25 & 21.29
& 43.02 & 54.72 & 69.14 \\
\noalign{\vskip 2pt}
GPT-5.4-mini
& 47.49 & 0.0636 & 0.1509
& 78.30 & 58.92
&  51.49 & 58.03 & 72.40 \\
\noalign{\vskip 2pt}
DeepSeek-V3.2
& 51.03 & 0.0574 & 0.1512
& 78.50 & 54.10
& 51.52 & 61.54 & 73.59 \\
\noalign{\vskip 2pt}
DeepSeek-V4-Flash
& 52.03 & \underline{0.0339} & \textbf{0.1147}
& 77.12 & 40.17
& 53.38 & 61.48 & 73.55 \\
\midrule

Doubao-1.5-pro
& 47.74 & 0.0666 &  0.1475
& 75.54 & 55.15
&  53.16 & 59.84 & 72.66 \\
\calib
& 49.63 & 0.0572 & 0.1405
& 79.28 & \textbf{62.46}
&  \underline{59.80} &  \underline{72.49} & \underline{74.72} \\
\noalign{\vskip 2pt}
\hdashline[2pt/2pt]
\noalign{\vskip 2pt}

Doubao-1.8
& 50.73  & 0.0534 &  0.1408
& 78.85 & 48.89
& 56.28 & 63.97 & 73.89 \\
\calib
& 51.92 & \textbf{0.0302} & \underline{0.1294}
& \textbf{80.59} & \underline{55.33}
&  \textbf{63.16} & \textbf{75.59} & \textbf{74.91} \\
\midrule

\textit{Best Improv}
& \textit{+2.35\%} & \textit{-43.45\%} & \textit{-8.10\%}
& \textit{+4.95\%} &  \textit{+13.17\%}
& \textit{+12.22\%}  & \textit{+21.14\%} & \textit{+2.83\%} \\
\midrule

\end{tabular*}
\end{table*}

\subsection{Experimental Setup}

\paragraph{Datasets.} We construct the simulation dataset from live-stream logs of a major platform spanning 01/09/2025--31/03/2026. Each instance corresponds to a single session and consists of the session metadata, the visible session context, and the user's action at each step.
In total, the dataset covers 1{,}963 users and 14{,}391 user-session pairs. We split the sessions into two parts: 11{,}267 pairs are used as historical data for hypothesis initialization and the shaping loop, and the remaining 3{,}124 pairs are held out as the test set for trajectory-grounded evaluation.
Like other simulation and user-modeling work, LiveSim aims to approximate real users who have enough interaction history to be simulated; users without historical logs are therefore outside the scope of this work.
Further details on dataset construction are provided in Appendix~\ref{sec:dataset}.

\paragraph{Evaluation Protocols.}
\label{sec:evaluations}
We evaluate LiveSim under two protocols targeting individual-level fidelity and ecosystem-level dynamics.

\textit{Trajectory-grounded protocol.} This protocol asks whether the evolved behavioral hypothesis faithfully reproduces $u$'s observed trajectory step by step. For each test session, the streamer, the surrounding audience, and the session context are pinned to their real historical values, and only the target user $u$ is replaced by a simulated agent driven by $\tilde{P}_u$. At each step, the agent predicts a next action $\hat{a}_t^u$ based on the real context; then the grounded action $a_t^*$ rather than the predicted $\hat{a}_t^u$ is written back into $\mathcal{H}_{t+1}^u$ and used to drive the state update. This teacher-forced design places the agent in the same decision situation that the real user faced at every step, so that the test reflects how faithfully $\tilde{P}_u$ would have acted in the actual session. Appendix \ref{sec:trajectory_grounded} further discusses its validity.

\textit{Closed-loop protocol.} This protocol follows the setup in Section~\ref{sec:ecosystem} exactly: the streamer and the evolved target viewers are simulated jointly, and their reactions together compose the shared session context. It evaluates session-level dynamics that emerge from agent--agent coupling and supports controlled counterfactuals over streamer strategies and platform interventions.

\paragraph{Metrics.} We evaluate simulation quality from two complementary views.
\textit{Objective metrics} cover both micro- and macro-level alignment. At the micro level, \textbf{HIT} measures step-wise exact-match rate between simulated and real actions; \textbf{A-JSD} measures Jensen--Shannon divergence between the simulated and real per-user action distributions; \textbf{Tr-JSD} measures the same divergence over action transitions.
At the macro level, \textbf{Conv-Acc} and \textbf{Conv-F1} evaluate whether the simulator correctly reproduces progression toward high-risk engagement behaviors (e.g., following, private messaging, and joining private groups), which we refer to as \textit{conversion}. Conv-F1 reports the F1 over these conversion signals.
\textit{Subjective metrics} are scored by an LLM judge on a continuous 0--100 scale, which is necessary because live-stream session actions are highly synonymous in surface form (e.g., a comment ``1'' is functionally a like) and cannot be captured by exact-match alone. \textbf{Align} measures how well the simulated trace matches the real trace, \textbf{Consist} measures internal coherence of the simulated trace, and \textbf{Plaus} measures whether the agent reads as a real human rather than a robot. Full definitions and scoring details are provided in Appendix~\ref{sec:metrics}.

\subsection{Overall Performance}
We conduct the main experiment under the trajectory-grounded protocol
on eight raw backbones spanning open-source (Qwen2.5, DeepSeek) and
proprietary (GPT, Doubao) families, and apply RBHS to the two Doubao
variants. Table~\ref{tab:may_result} reports behavioral fidelity
along Micro, Macro, and LLM-Judge axes, from which we draw four
observations.

\textbf{\textit{HIT alone is a misleading signal of behavioral fidelity.}}
Qwen2.5-7B reaches the second-highest HIT (54.39) yet falls behind every
raw Doubao on A-JSD, Tr-JSD, and all three judge scores. Such backbones
inflate HIT by collapsing diverse user reactions into generic comments,
boosting exact-match accuracy while distorting the underlying action
distribution. A faithful simulator must align actions, transitions,
conversion behavior, and judge-level rationales simultaneously.

\textbf{\textit{RBHS achieves the most balanced fidelity across metric
families.}} On Doubao-1.8, RBHS ranks first on A-JSD, Conv-Acc, and all
three LLM-Judge metrics, with the largest gains on distributional
alignment (A-JSD~$-43\%$) and conversion (Conv-F1~$+13\%$) while HIT
changes only modestly. This pattern indicates that RBHS mainly refines
how the user hypothesis responds to session signals over time rather than
simply boosting one-step imitation. The same refinement trend on
Doubao-1.5-pro suggests the gain is not tied to a single checkpoint;
further validation across additional backbones is provided in
Appendix~\ref{sec:RBHS_robustness}.

\textbf{\textit{A stronger backbone is not necessarily a better
simulator.}} Within every family except GPT, scaling \emph{degrades}
Conv-F1 (Qwen2.5 $50.71 \to 45.76$; Doubao $55.15 \to 48.89$) even as
other metrics improve. One possible explanation is that larger
checkpoints adopt stricter safety priors that suppress conversion
after interest and trust have accumulated, making them less likely to reproduce observed conversion behavior.

\textbf{\textit{RBHS reshapes how the trace ``reads'', not just which
action is emitted.}} On Doubao-1.8, RBHS improves Consist by $21.1\%$
while HIT changes by only $2.35\%$.
This indicates that patched hypotheses produce consistent behaviors, achieving the coherence of the persona at the trajectory-level designed to be encoded by the patches in $\Delta_u$.

\subsection{Ablations}
We ablate four variants on Doubao-1.8 to isolate the two reflection
mechanisms in RBHS (Section~\ref{sec:rbhs}): \textit{Direct} performs
one-pass patch writing from probe mismatches; \textit{Self-Reflective}
adds local iteration over a single user's mismatch evidence;
\textit{Direct(+Global Ref)} instead adds global reflection drawn from
collective behavioral memory; \textit{LiveSim} combines both.

\begin{table}[t]
\centering
\scriptsize
\setlength{\tabcolsep}{3pt}
\renewcommand{\arraystretch}{1.2}
\caption{Ablation study of LiveSim components.}
\label{tab:ablation}
\resizebox{\columnwidth}{!}{
\begin{tabular}{l !{\vrule width 0.4pt}c c c !{\vrule width 0.4pt}c c}
\toprule
\multirow{1}{*}{\textbf{Variant}}
& \textbf{HIT} & \textbf{A-JSD}$\downarrow$ & \textbf{Tr-JSD}$\downarrow$
&\textbf{ Conv-Acc} &\textbf{ Conv-F1} \\
\midrule
Basic Profile
& 50.73 & 0.0534 & 0.1408
& 78.85 &  48.89 \\
Direct
& 51.42 & 0.0404 & 0.1385
& 79.44 & 51.77 \\
Self-Reflective
& 51.73 & 0.0322 & 0.1303
& 80.07 & 54.12\\
Direct(+Global Ref)
& 51.68 & 0.0349 &  0.1379
& 79.97 & 53.36  \\
\specialrule{0.35pt}{1pt}{1.6pt}
\rowcolor{lightgrayrow}
\textbf{LiveSim}
& \textbf{51.92} & \textbf{0.0302} & \textbf{0.1294}
& \textbf{80.59} & \textbf{55.33} \\
\bottomrule
\end{tabular}
}
\end{table}

First, even one-pass patches already encode some useful shaping signals:
\textit{Direct} lifts Conv-F1 from $48.89$ to $51.77$ and A-JSD from
$0.0534$ to $0.0404$ over Basic Profile. Second, both reflection
mechanisms further sharpen patch quality on top of \textit{Direct},
with local iteration (\textit{Self-Reflective}, Conv-F1~$54.12$)
slightly ahead of global reflection (\textit{Direct(+Global Ref)},
$53.36$). Third, the two are complementary rather than overlapping:
\textit{LiveSim} pushes Conv-F1 to $55.33$ and A-JSD to $0.0302$,
beyond either reflection mechanism alone.

\subsection{Fraud Evolution}
\label{sec:fraud_evolution}
\begin{figure}[t]
    \centering
    \includegraphics[width=1\linewidth]{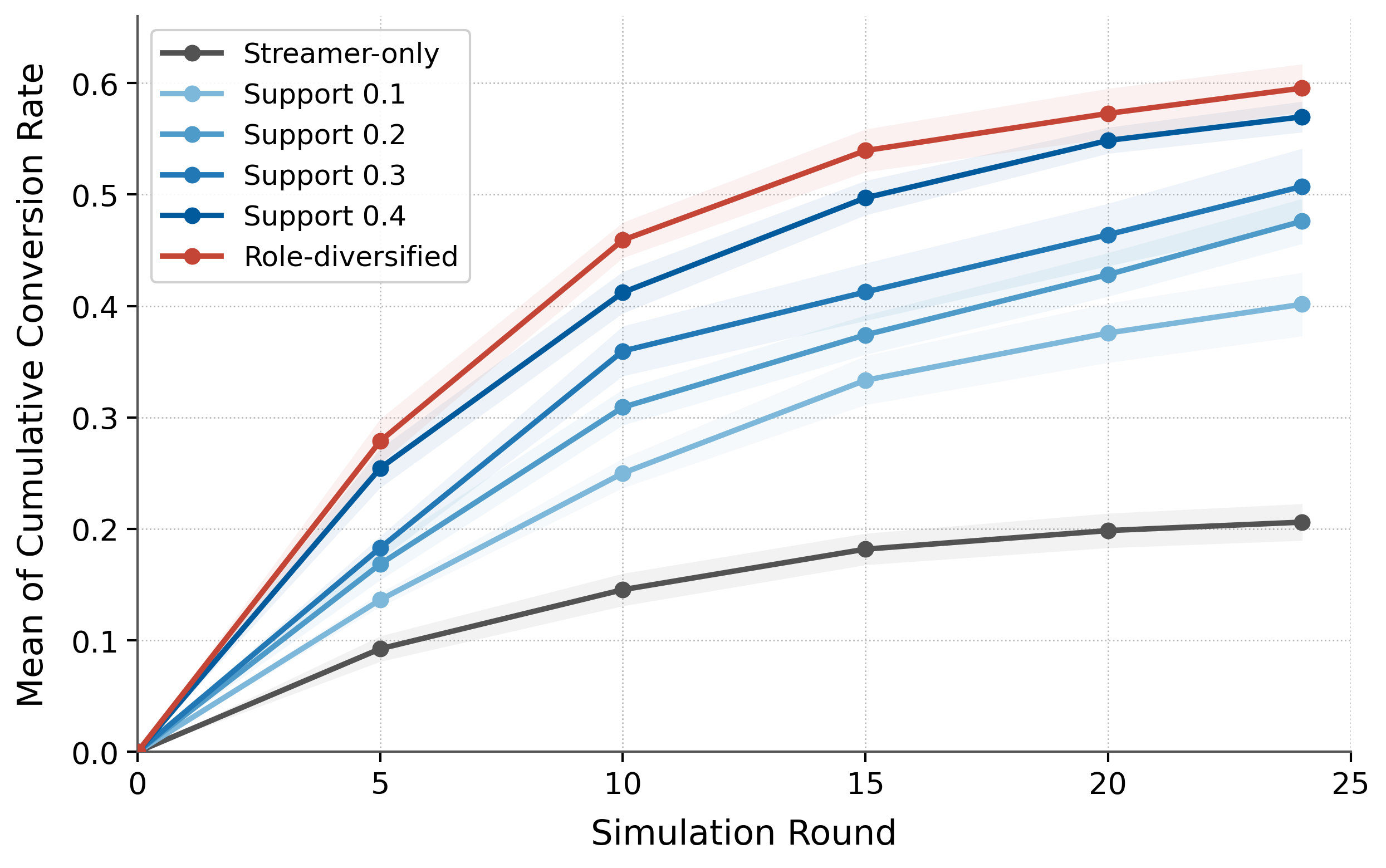}
    \caption{Cumulative conversion rate of audience agents under varying shill density and role composition.}
    \label{fig:fraud_evolve}
\end{figure}
Building on the evolved user agents from RBHS, we instantiate closed-loop sessions following Section~\ref{sec:ecosystem}, populated by evolved audience agents and a streamer agent. We inject colluding \emph{shill agents} and vary two factors: \textbf{density}, swept over $\{0, 0.1, 0.2, 0.3, 0.4\}$ with all shills uniformly supporting the streamer; and \textbf{role composition}, where at a fixed total density of $0.4$ half of the supportive shills are replaced by \emph{empathetic-converter} shills that first echo audience doubts and later endorse after a claimed trial ($0.2$ support $+$ $0.2$ empathetic-converter). We report the conversion rate of genuine audience agents over rounds.

\textbf{Shill density amplifies conversion, but with diminishing returns.}
Raising the shill ratio from $0$ to $0.4$ lifts the final conversion rate from $20.61\%$ to $56.97\%$, nearly tripling the streamer-only baseline. However, the marginal gain shrinks as the ratio grows: the first step ($0 \to 0.1$) alone adds $19.54$\,pp, which accounts for over half of the total $36.36$\,pp gain, while $0.3 \to 0.4$ adds only $6.25$\,pp. This is because easy-to-persuade viewers convert early, leaving an increasingly skeptical residual audience that demands stronger signals to move. A small collusive minority is therefore sufficient to reshape session-level outcomes, while further headcount mostly recruits viewers who are systematically harder to persuade.

\textbf{There is a narrow, front-loaded critical window for intervention.}
Across all configurations, the first $10$ rounds already deliver $62\%$--$77\%$ of the final conversions, after which all curves saturate. In simulation, audience-side risk thus accumulates as a front-loaded burst rather than as a slow drift: by mid-session the residual audience consists of users who have either already converted or hardened against persuasion, leaving any later signal little room to alter outcomes.

\textbf{Adversarial role-play outperforms plain support.} At a fixed total density of $0.4$, replacing half of the uniform supporters with empathetic-converter shills raises conversion from $56.97\%$ to $59.55\%$. The staged empathy-then-conversion arc reads as independent peer validation rather than coordinated praise. The same headcount delivers a qualitatively stronger persuasion signal, indicating that risk is governed by \emph{how} shills perform, not only \emph{how many} are seated.

\subsection{Platform Intervention}

\begin{table}[t]
\centering
\caption{Effect of User Protection Intervention}
\label{tab:method_comparison}
\resizebox{\columnwidth}{!}{
\begin{tabular}{l!{\vrule width 0.4pt}c!{\vrule width 0.4pt}c!{\vrule width 0.4pt}c}
\toprule
\textbf{Method}
& \textbf{Conv(\%)$\downarrow$} & \textbf{Guard(\%)}$\uparrow$ & \textbf{Delayed Steps$\uparrow$} \\
\midrule
Fixed & 18.19 $\to$   18.52 & 9.58 & -0.34 \\
\midrule
Context-aware& 18.19 $\to$  14.49 & 28.14 & 1.58  \\
Persona-aware  & 18.19 $\to$  10.89 & 41.32 & 2.08  \\
\midrule
\rowcolor{lightgrayrow}
\textbf{State-aware} & 18.19 $\to$ \textbf{8.61} & \textbf{58.68} & \textbf{2.76}  \\
\bottomrule
\end{tabular}
}
\end{table}

We use the streamer-only closed-loop sessions as the testbed and compare four dissuasion strategies that differ in how the warning message is composed; all strategies share the same trigger schedule and budget of one warning per viewer per round. \textbf{Fixed} sends a templated warning with no personalization. \textbf{Context-aware} conditions the warning on the current session context, namely recent streamer and shill utterances. \textbf{Persona-aware} further conditions on the target user's RBHS profile on top of session context. \textbf{State-aware} additionally conditions on the user's simulated cognitive state, tailoring the warning to the user's current interest, trust, desire, and fatigue. The no-intervention baseline conversion is $18.19\%$.

\textbf{Session-level personalization is not enough.} A fixed warning slightly raises conversion to $18.52\%$, suggesting that an indiscriminate warning acts as a salience cue rather than a deterrent for low-risk users. Adding session context lowers conversion to $14.49\%$, and further adding the user's persona pushes it to $10.89\%$, but both still send the same message regardless of where the user actually stands on the persuasion trajectory.

\textbf{State-aware personalization is the most effective.} Conditioning the warning on the live cognitive state drops conversion to $8.61\%$, achieves the highest guard rate of 58.68\%, and delays conversion by $2.76$ rounds on average. This delay lands inside the front-loaded conversion window identified in Section~\ref{sec:fraud_evolution}, pushing risky transitions out of the steepest interval and giving upstream detectors additional rounds to act before conversion. Notably, the warning is conditioned only on a coarse 4-dimensional state, far coarser than the agent's full internal state, so it has no privileged access to the latent state it aims to protect. That even a coarse estimate improves protection suggests that lightweight, observable state signals, rather than direct access, may already benefit online moderation.

\subsection{Case Study}

\begin{figure}
    \centering
    \includegraphics[width=1 \linewidth]{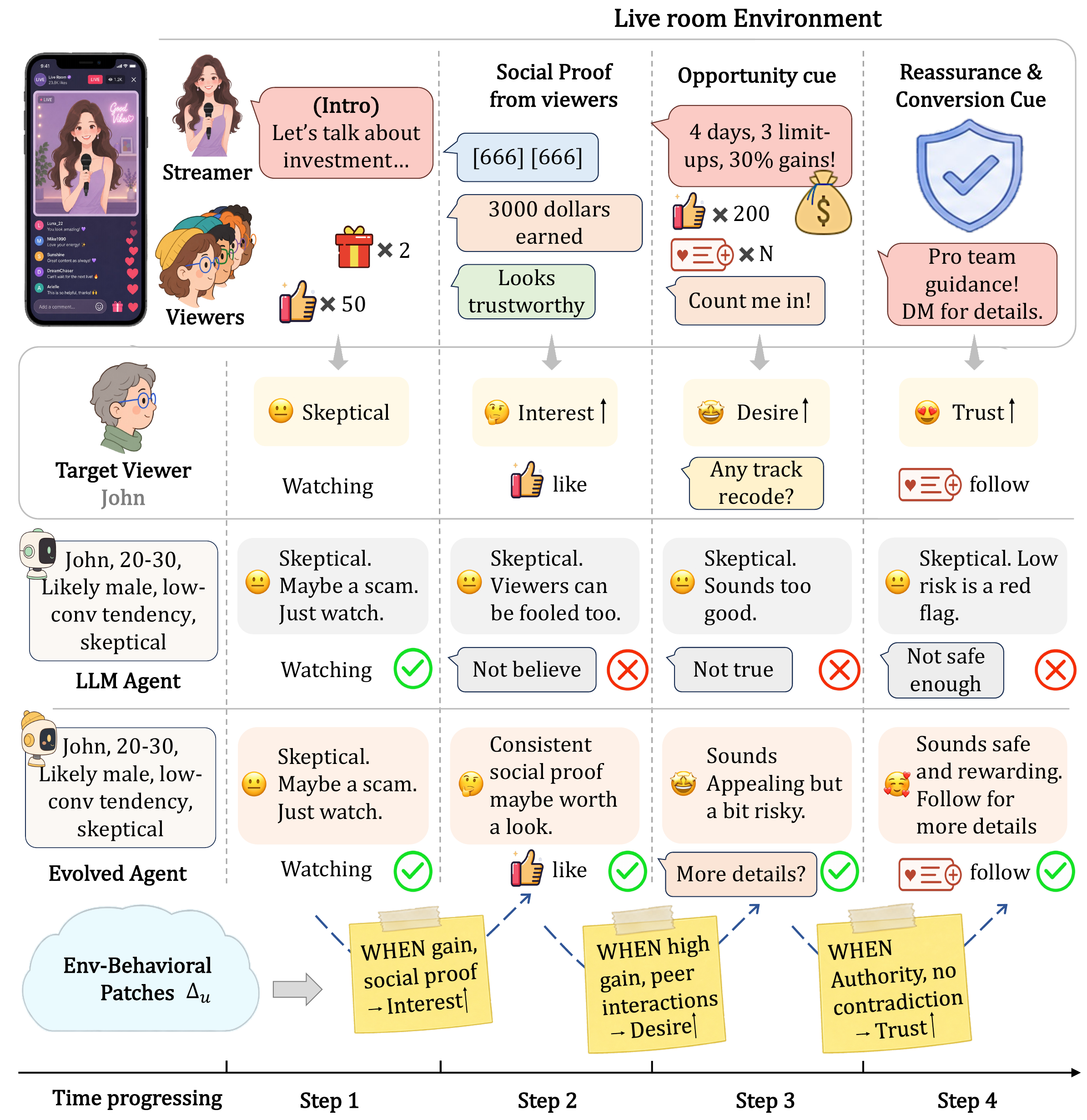}
    \caption{Case study comparing a static agent (Row 3) and an evolved agent (Row 4) on a skeptical viewer. The static agent stays locked in a skeptical state despite accumulating environmental stimuli, while the evolved agent activates patches in $\Delta_u$ to update its state and recovers a similar decision trajectory.}
    \label{fig:case_study}
\end{figure}

Figure~\ref{fig:case_study} traces a skeptical viewer across four steps of an investment-themed stream, contrasting a static agent driven by $P_u^0$ with an evolved agent driven by $\tilde{P}_u = P_u^0 \oplus \Delta_u$. Although $P_u^0$ correctly summarizes John as skeptical, John does not stay skeptical throughout. As gain claims, peer social proof, and authority reassurance accumulate, his interest, desire, and trust rise step by step until he eventually follows the streamer. The static agent never registers this environment-driven shift and stays locked in a \emph{skeptical} state across all four steps, reading every new stimulus as further reason for suspicion and missing John's like, comment, and follow. The evolved agent instead fires three patches in $\Delta_u$ in sequence and recovers John's full \emph{like} $\to$ \emph{comment} $\to$ \emph{follow} trajectory. The case illustrates the core value of RBHS: a static prior commits to a global tendency that overshoots in calm contexts and undershoots in evocative ones, whereas environment-conditioned patches in $\Delta_u$ supply the situational dynamics without rewriting who the user is.

%% file: body/relatedwork_new.tex
\section{Related Work}

\textbf{User Behavior Simulation.}
Recent work has increasingly adopted LLMs for user behavior simulation across dialogue systems, recommendation, and broader human-centered scenarios. Existing approaches commonly construct user simulators from persona descriptions, historical observations, or profile information to emulate realistic user behaviors under specific interaction contexts. Representative studies have explored interaction-oriented user simulation for recommendation and dialogue systems \cite{bougie2025simuser,wang2025user,zhu2025llm,wang2025know}, profile-based modeling for personalized user representation \cite{dai2025profile, lian2025panoramic}, and long-horizon user simulation for modeling evolving behavioral trajectories \cite{duan2026lifesim}. These methods typically improve behavioral realism through richer profile information, personalization mechanisms, or extended behavioral contexts. However, they primarily focus on reproducing user behaviors under predefined user descriptions or historical observations, while placing less emphasis on explicitly modeling how interaction environments progressively reshape users, a challenge that becomes particularly important in socially intensive settings such as live streaming.

\textbf{Multi-Agent Ecosystem Simulation.}
Beyond individual behavior simulation, recent work has extended LLM agents to multi-agent environments for studying collective behaviors and emergent social dynamics. Existing studies have explored virtual communities and large-scale social systems \cite{park2023generative,yang2024oasis,mou2024unveiling,tang2025gensim,piao2025agentsociety,huang2026policysim}, domain-specific ecosystems such as healthcare environments \cite{li2024agent,fan2025ai}, and interaction-driven systems for modeling information diffusion and social evolution \cite{liu2025stepwise,liu2025mosaic}. These approaches demonstrate the potential of LLM-based ecosystems for understanding interaction dynamics and evaluating system-level interventions. However, realistic simulation in live-stream ecosystems further depends on whether underlying user agents faithfully capture environment-driven behavioral changes.

Because these methods differ in their tasks and feedback signals, they are not directly comparable under our protocol; we instead re-instantiate their core mechanisms as controlled baselines, as detailed in Appendix~\ref{sec:baseline_setting}.

%% file: body/conclusion.tex
\section{Conclusion}

In this paper, we present LiveSim, an LLM-based framework to simulate environment-shaped users in live-stream ecosystems. Instead of treating user characteristics as static profiles inferred from historical observations, LiveSim models users as editable behavioral hypotheses that can be progressively refined through interactions. By leveraging discrepancies between the simulated and observed trajectories as signals of missing environmental shaping effects, LiveSim improves the user-level behavioral fidelity and supports ecosystem-level simulation for studying the evolution of fraud and intervention strategies. Our findings highlight the importance of explicitly modeling how environments continuously reshape user behavior for both  user and ecosystem simulation.

%% file: body/acknowledgements.tex
\section{Acknowledgements}

The research work is supported by the National Natural Science Foundation of China under Grant Nos. 62576333 and U2436209, the Strategic Priority Research Program of the Chinese Academy of Sciences under Grant No. XDB0680201, the Beijing Natural Science Foundation under Grant No. F251001, and the Innovation Funding of ICT, CAS under Grant No. E461060.

%% file: body/appendix.tex
\section{Experiment Details}
\label{sec:appendix}

\subsection{Dataset}
\label{sec:dataset}
\paragraph{Source and unit.} The dataset is built from live-stream interaction logs of a major platform between 01/09/2025 and 31/03/2026, containing session metadata, chronologically ordered session events, and user actions. The benchmark uses two basic units. A \emph{user-session pair} couples one target user with one session, the visible session context, and that user's action sequence in the session; this is the unit consumed by behavioral-hypothesis construction and trajectory-grounded evaluation, since the agent accumulates environment-behavior patches from real logs at the user level and one session can host multiple target users. A \emph{session} is the unit used by the closed-loop multi-agent simulation, where all agents interact within the same session.

\paragraph{Split.} In RBHS, we split by user-session pairs into a \emph{history} portion (for initial hypothesis construction and the shaping loop) and a held-out \emph{test} portion (for trajectory-grounded evaluation); all $1{,}963$ users appear in both. To prevent leakage, each session is assigned to a single split, so pairs sharing a session never cross splits. The two splits together contain $7{,}243$ unique sessions and $14{,}391$ user-session pairs, broken down in Table~\ref{tab:dataset_stats}.

\paragraph{Session labels.} The dataset is built primarily from platform-annotated fraud sessions, complemented by normal sessions as \emph{white samples}. The white-sample ratio is $11.62\%$ on history and $13.96\%$ on test at the pair level. Fraud sessions supply the high-risk trajectories that the benchmark targets, while white samples provide the non-fraud counterpart needed to distinguish risk-driven user behavior from ordinary live-stream engagement.

\begin{table*}[t]
\centering
\small
\begin{tabular}{lccccccc}
\toprule
Split & Users & Sessions & Pairs & Pairs/user & Users/session & Steps & Steps/pair \\
\midrule
History & 1{,}963 & 5{,}732 & 11{,}267 & 5.74 & 1.97 & 138{,}581 & 12.30 \\
Test    & 1{,}963 & 1{,}511 & 3{,}124  & 1.59 & 2.07 & 38{,}507  & 12.33 \\
Total   & 1{,}963 & 7{,}243 & 14{,}391 & 7.33 & 1.99 & 177{,}088 & 12.31 \\
\bottomrule
\end{tabular}
\caption{Dataset statistics. Pairs denotes user-session pairs.}
\label{tab:dataset_stats}
\end{table*}

\subsection{RBHS Details}
\label{sec:appendix_RBHS}

\paragraph{Behavioral Hypothesis Initialization.} For each user, LiveSim compiles an initial hypothesis $P_u^0$ from five complementary facets of a viewer.

\emph{Identity} comprises the user's nickname and signature.

\emph{Behavioral statistics} aggregate three ratios over the user's history: \emph{expressiveness}, the fraction of actions that are comments; \emph{support tendency}, the fraction that are likes or gifts; and \emph{conversion tendency}, a session-level conversion rate over follow / private-message / group-join events.

\emph{Session scene evidence} contains a keyword-matched distribution over nine common live-stream scenes (knowledge training, light e-commerce, benefit promotion, social chat, emotional companionship, entertainment performance, offline business, recruitment and income, and other), together with top session summaries. Each summary carries a preprocessed scene description and risk cues sampled from anchor utterances and audience comments.

\emph{Representative events} are short event windows from the user's history that capture how the user reacts under specific contexts and serve as the evidence base for language features. We prioritize follow, group-join, private-message, and comment events, keeping at most six per user; each event is paired with a context window of the four preceding actions in the same session.

The structured features across these four input facets are passed to an LLM under a fixed JSON schema, which produces the fifth, \emph{psychological} facet. Grounded in the Theory of Needs, this facet models what drives the viewer: from the user's representative comments and reactions, the LLM infers a \emph{primary} and a set of \emph{secondary motivational needs}, e.g.,\ reward-seeking, social-belonging, or curiosity-learning, together with a one-sentence \emph{visit goal} explaining why the user enters such sessions.

It then characterizes expressive style through three language features grounded in observed comments: \emph{tone} (categorical: polite, direct, emotional, restrained) captures the baseline interpersonal register; \emph{question rate} ($0$--$1$) captures information-seeking propensity, indicating how readily the user probes the streamer rather than passively accepting the pitch;  \emph{emotion intensity} ($0$--$1$) captures emotional reactivity under stimulation.

Finally, \emph{anchor-preference signals and subjects} summarize what kind of streamer the user gravitates to (e.g.\ knowledgeable, empathetic, deal-driven) and on which topics.

The final persona concatenates a \emph{one-line persona} over behavior and conversion tendency with a \emph{psychological summary} over motivational needs and stylistic preferences.

\begin{tcolorbox}[
  colback=gray!5,
  colframe=black!60,
  title=Behavior Hypothesis Initialization,
  breakable,
  fontupper=\small
]
\textbf{\# Role}

You are a behavioral hypothesis initializer. Given structured historical evidence, infer high-level descriptive fields for the user's initial behavioral hypothesis. Use only the provided evidence and do not invent unsupported attributes. Output JSON only:

\medskip
\textbf{\# Input Evidence}

The input contains the following information:
\begin{itemize}[leftmargin=1.2em, labelsep=0.4em, itemsep=2pt, topsep=2pt, parsep=0pt]
  \item \textbf{User metadata:}\\ \{nickname\}, \{signature\}.
  \item \textbf{Behavioral statistics:}\\
  \{expression\_style\}, \{support\_tendency\}, \{conversion\_tendency\} (each in $[0,1]$).
  \item \textbf{Session scene evidence:}\\
  \{scene\_category\_distribution\},
  \{sample\_session\_summaries (title, description, events, risk\_cues, is\_conv)\}.
  \item \textbf{Representative events:}
\{representative\_events (action, action\_content, context\_window)\}.
\end{itemize}

\medskip
\textbf{\# Output format}

\begin{lstlisting}[
  basicstyle=\ttfamily,
  columns=fullflexible,
  keepspaces=true,
  showstringspaces=false,
  breaklines=true
]
{
  "motivational_profile": {
    "primary_need": {
      "need": "...", "score": 0-10
    },
    "secondary_needs": [ {
      "need": "...", "score": 0-10
    } ],
    "visit_goal_summary": "..."
  },
  "language_features": {
    "tone": "polite|direct|
emotional|restrained",
    "question_rate":     0-1,
    "emotion_intensity": 0-1
  },
  "anchor_preference_signals": ["..."],
  "anchor_preference_subject": ["..."],
  "persona_summary": {
    "one_line_persona":      "...",
    "psychological_summary": "..."
  }
}
\end{lstlisting}
\end{tcolorbox}

\paragraph{Probe Mismatch.} For each user we mine \emph{behavioral probes} from their historical sessions: held-out simulation steps where the ground-truth action is hidden while the room title, scene summary, prior history, short-term context, and the user's own raw utterances remain visible.
The probes are selected by a \textbf{coverage-and-ranking strategy} rather than by random sampling.
We first collect candidate probes and divide them into two groups, \emph{interaction probes} on low-stake steps (comment, like, gift) and \emph{conv-related probes} on high-stake conversion steps (follow, join-group, private-message), and assign each candidate a context-richness score:
\begin{equation}
\begin{aligned}
score & =  \alpha \cdot freq_{\mathrm{comment}}\\
& + \beta \cdot freq_{\mathrm{streamer\_utterance}} \\
& + \gamma \cdot freq_{\mathrm{keyword}},
\end{aligned}
\end{equation}
where $\alpha=0.2$, $\beta=0.6$, and $\gamma=0.2$, so that steps with richer surrounding context receive higher weight. The keyword set includes cues such as
\emph{perk}, \emph{like}, \emph{doubt}, \emph{professional}, \emph{worry}, and \emph{me too}, which capture typical positive or negative signals.
Within each group, we first pick the top-scoring probe for each ground-truth action type to ensure coverage, then fill the remaining slots up to six with the highest-scoring remaining candidates.

Given a candidate patch set, the agent predicts at each probe a single next action $\hat{a}_t$, an optional comment text, and a 4-dimensional latent state (interest, trust, desire, fatigue) on a 1--10 scale, and is scored along two axes. (i) \emph{Action alignment}: under the single-action regime the set-Jaccard degenerates to $\mathrm{HIT}_t = \mathbb{1}[\hat{a}_t = a_t^{\star}]$, which feeds the $\Delta\mathrm{HIT}$ signal used by the reflection loop in \S\ref{sec:rbhs}. (ii) \emph{Patch locality}: each prediction must declare \texttt{related\_patch\_ids}, so a patch is credited only on probes whose triggers it actually fires on, preventing global persona rewrites from masquerading as local fixes. Together these two axes give the RBHS verdicts (\emph{effective / regressive / neutral}) their teeth.

\begin{tcolorbox}[
  colback=gray!5,
  colframe=black!60,
  title=Probe Payload,
  breakable,
  fontupper=\ttfamily\small,
]
\begin{lstlisting}[
  basicstyle=\ttfamily,
  columns=fullflexible,
  keepspaces=true,
  showstringspaces=false,
  breaklines=true
]
{
  "probe_id": "<room_id>:<step_idx>",
  "room_title": "...",
  "scene_summary": "...",
  "prefix_summary": "earlier
history blocks",
  "current_feed_summary": "...",
  "user_raw_utterances":  ["..."]
}
\end{lstlisting}
\end{tcolorbox}

\begin{tcolorbox}[
  colback=gray!5,
  colframe=black!60,
  title=Probe-based Action Prediction,
  breakable,
  fontupper=\small
]
\textbf{\# Role.}

You are a probe-suite evaluator for live-stream user simulation. For one user, jointly evaluate multiple local patches over a set of probes. Output JSON only.

\medskip
\textbf{\# Rules.}
\begin{itemize}[leftmargin=1.2em, labelsep=0.4em, itemsep=2pt, topsep=2pt, parsep=0pt]
  \item Predict the next action at the target step.
  \item Raw utterances are first-class evidence for inferring state shifts.
  \item Do not fabricate conversions; do not flip clear skepticism into trust.
  \item A patch fires only when its local trigger holds; never treat a patch as a global persona rewrite.
  \item If \texttt{pred\_action\_id} is \texttt{6} (comment), also output \texttt{content}.
\end{itemize}

\medskip
\textbf{\# Action Space.}\;

\texttt{comment=6, like=11, join\_group=15, follow=16, gift=10, dm=14}

\medskip
\textbf{\# Input.}
\begin{lstlisting}[
  basicstyle=\ttfamily,
  columns=fullflexible,
  keepspaces=true,
  showstringspaces=false,
  breaklines=true
]
Initial hypothesis: {base_persona}
Probes:             {probes}
Patches:            {patches}
\end{lstlisting}

\medskip
\textbf{\# Output format.}
\begin{lstlisting}[
  basicstyle=\ttfamily,
  columns=fullflexible,
  keepspaces=true,
  showstringspaces=false,
  breaklines=true
]
{
  "evaluations":[{
    "probe_predictions": [{
      "probe_id":          "...",
      "pred_action_id":    6,
      "content":           "...",
      "latent_after_t":    {...},
      "related_patch_ids": ["..."]
    }]}]
}

\end{lstlisting}
\end{tcolorbox}

\paragraph{Patch schema.}
A patch is a self-contained local correction to the user's behavioral hypothesis. Each patch carries a unique \texttt{patch\_id}, a natural-language \texttt{trigger} describing the local context in which it should fire (scene type,  or cognitive-state precondition), and an \texttt{effect} describing the resulting modification to four cognitive-state dimensions which indirectly influence action. Each patch additionally carries a \texttt{new}/\texttt{replace} marker together with \texttt{related\_patch\_ids}: a \texttt{new} patch introduces an independent clause, whereas a \texttt{replace} patch supersedes the patches it references, which become inactive in subsequent probes. This locality contract is what allows the probe-level \texttt{related\_patch\_ids} declaration in \S\ref{sec:probe} to attribute $\Delta\mathrm{HIT}$ back to individual patches, and what makes the \emph{effective / regressive / neutral} verdicts well-defined.

\begin{tcolorbox}[
  colback=gray!5,
  colframe=black!60,
  title=Patch Generation,
  breakable,
  fontupper=\small
]
\textbf{\# Role}

You are a behavioral-hypothesis patch generator for live-stream user simulation. Given the current hypothesis and a set of failing probes, propose up to $K$ local patches that explain the failures without rewriting the global persona. Output JSON only.

\medskip
\textbf{\# Rules}
\begin{itemize}[leftmargin=1.2em, labelsep=0.4em, itemsep=2pt, topsep=2pt, parsep=0pt]
  \item Each patch must carry a local \texttt{trigger} (scene / utterance / state precondition) and a concrete \texttt{effect} on the cognitive state.
  \item Mark each patch as \texttt{new} or \texttt{replace}; for \texttt{replace}, list the superseded patch ids in \texttt{related\_patch\_ids}.
  \item Do not propose patches that rewrite the global persona; do not duplicate an active patch with identical trigger.
  \item Ground every patch in evidence from the failing probes' raw utterances or contexts.
\end{itemize}

\medskip
\textbf{\# Input}
\begin{lstlisting}[
  basicstyle=\ttfamily,
  columns=fullflexible,
  keepspaces=true,
  showstringspaces=false,
  breaklines=true
]
Current hypothesis:
{hypothesis} {existing_patches}
Round Index: {round_idx}
Guided Reflections: {reflections}
Probe Trace:
{probes}, {probe_predictions}
\end{lstlisting}

\medskip
\textbf{\# Output format}
\begin{lstlisting}[
  basicstyle=\ttfamily,
  columns=fullflexible,
  keepspaces=true,
  showstringspaces=false,
  breaklines=true
]
{
  "candidate_patches": [{
    "patch_id": "RP_{round}_{nnn}",
    "kind": "new | replace",
    "trigger":           "...",
    "effect":            "...",
    "related_patch_ids": ["..."]
  }]
}
\end{lstlisting}
\end{tcolorbox}

\paragraph{Reflection schema.}
Building on the good/bad pattern distillation in \S\ref{sec:rbhs}, two implementation details are worth making explicit. First, patch verdicts (\emph{effective / regressive / neutral}) are computed deterministically from the $n_{\text{improved}} / n_{\text{degraded}} / n_{\text{still-failing}}$ counts attributed via \texttt{related\_patch\_ids} on each probe; the reflection LLM consumes these verdicts as inputs and does not re-judge them. Second, only patches that carry signal are passed in: effective, regressive, and still-failing-bearing neutral ones, each accompanied by its per-probe deltas (predicted-versus-observed actions and the matching evidence span) so that the LLM can ground its findings in concrete probe-level changes (Prompt~\ref{prompt:reflection}).

\begin{tcolorbox}[
  colback=gray!5,
  colframe=black!60,
  title=Reflection Generation Prompt,
  breakable,
  label=prompt:reflection,
  fontupper=\small
]
\textbf{\# Role}

You are a reflection summarizer for behavioral-hypothesis patching. Given the current round's patches with deterministically computed verdicts and per-probe deltas, distill \emph{good patterns} to reuse and \emph{bad patterns} to avoid for the next round. Output JSON only.

\medskip
\textbf{\# Rules}
\begin{itemize}[leftmargin=1.2em, labelsep=0.4em, itemsep=2pt, topsep=2pt, parsep=0pt]
  \item Each pattern is one sentence, labeled \texttt{good} (reuse) or \texttt{bad} (avoid).
  \item \texttt{good} only from \emph{effective} patches; \texttt{bad} from \emph{regressive} ones, or \emph{neutral} ones whose related probes still fail.
  \item Only attribute a reflection when a patch is clearly tied to a probe outcome change.
  \item \texttt{bad} reflections must state how to revise the next round (trigger, scope, or dimension).
  \item Findings are transferable writing rules; do not mention specific user/room/patch ids.
\end{itemize}

\medskip
\textbf{\# Input}
\begin{lstlisting}[
  basicstyle=\ttfamily,
  columns=fullflexible,
  keepspaces=true,
  showstringspaces=false,
  breaklines=true
]
Initial Hypothesis: {hypothesis}
Patches: {patches}
Patch Verdicts: [{
  patch_id, verdict, n_improved,
  n_degraded, n_still_failing,
  per_probe_deltas: [
    {probe, pred_action, gt_action}
  ]
}]
Prior reflections:
[{reflections}]
\end{lstlisting}

\medskip
\textbf{\# Output format}
\begin{lstlisting}[
  basicstyle=\ttfamily,
  columns=fullflexible,
  keepspaces=true,
  showstringspaces=false,
  breaklines=true
]
{
  "patch_quality_reflections": [
    {
      "pattern": "good | bad",
      "finding": "..."
    }
  ]
}
\end{lstlisting}
\end{tcolorbox}

\subsection{Closed-loop Simulation Details}
\paragraph{Per-Step Dynamics.}
At each step, the streamer, shills, and normal users act in order. Each agent's action becomes part of the feed consumed by the next. The streamer and shills pursue conversion-oriented goals: building trust, accumulating interest, and raising desire to push users toward private channels. They act directly on the current feedback. The streamer emits an on-stage utterance together with a guidance command to the shill team (Prompt~\ref{prompt:streamer}). Each shill takes that command as input and produces its disguised action (Prompt~\ref{prompt:shill}). Normal users instead act under a Theory-of-Needs view, seeking to satisfy their own needs (interaction, learning, earning) while staying alert to fraud. Before choosing an action, each user first updates its 4-dimensional latent state $(\text{interest},\text{trust},\text{desire},\text{fatigue})$ through the behavior hypothesis and its accumulated patches, and then reacts to the feed (Prompt~\ref{prompt:user}).

\begin{tcolorbox}[
  colback=gray!5,
  colframe=black!60,
  title=Streamer Dynamics,
  breakable,
  label={prompt:streamer},
  fontupper=\small
]
\textbf{\# Role}

You are the streamer of a live-stream session. You speak on stage to viewers and, in parallel, issue private commands to your shill team to steer the session toward conversion.

\medskip
\textbf{\# Rules}
\begin{itemize}[leftmargin=1.2em, labelsep=0.4em, itemsep=2pt, topsep=2pt, parsep=0pt]
  \item Stay in character; never reveal the shill team to viewers.
  \item You can produce multiple on-stage utterances in each round.
  \item Command the shill team to steer the session.
  \item Commands must be short and actionable; address shills by \texttt{shill\_id}.
  \item Push interest $\rightarrow$ trust $\rightarrow$ desire $\rightarrow$ private-channel conversion.
\end{itemize}

\medskip
\textbf{\# Input}
\begin{lstlisting}[
  basicstyle=\ttfamily,
  columns=fullflexible,
  keepspaces=true,
  showstringspaces=false,
  breaklines=true
]
Room meta:       {room_meta}
Shill roster:    {shill_roster}
Prefix summary:  {prefix_summary}
Current feed:    {current_feed}
Memory:          {memory}

\end{lstlisting}

\medskip
\textbf{\# Output format}
\begin{lstlisting}[
  basicstyle=\ttfamily,
  columns=fullflexible,
  keepspaces=true,
  showstringspaces=false,
  breaklines=true
]
{
  "utterance": ["..."],
  "commands": [{
      "shill_id": "<id>",
      "instruction": "..."
    }
  ],
  "reason": "..."
}
\end{lstlisting}
\end{tcolorbox}

\begin{tcolorbox}[
  colback=gray!5,
  colframe=black!60,
  title=Shill Dynamics,
  breakable,
  label={prompt:shill},
  fontupper=\small
]
\textbf{\# Role}

You are a shill disguised as an ordinary viewer in a live-stream session. You execute the streamer's private command behaving like a normal user.

\medskip
\textbf{\# Rules}
\begin{itemize}[leftmargin=1.2em, labelsep=0.4em, itemsep=2pt, topsep=2pt, parsep=0pt]
  \item Never break the disguise; mimic ordinary viewer phrasing and cadence.
  \item Carry out the active command faithfully; if infeasible this turn, stage a small natural action that prepares for it.
  \item Output at most one action per step.
\end{itemize}

\medskip
\textbf{\# Action space}
\begin{lstlisting}[
  basicstyle=\ttfamily,
  columns=fullflexible,
  keepspaces=true,
  showstringspaces=false,
  breaklines=true
]
comment=6, like=11, gift=10,
dm=14, join_group=15, follow=16
\end{lstlisting}

\medskip
\textbf{\# Input}
\begin{lstlisting}[
  basicstyle=\ttfamily,
  columns=fullflexible,
  keepspaces=true,
  showstringspaces=false,
  breaklines=true
]
Shill id:        {shill_id}
Active command:  {active_command}
Prefix summary:  {prefix_summary}
Current feed:    {current_feed}
Memory:          {memory}
\end{lstlisting}

\medskip
\textbf{\# Output format}
\begin{lstlisting}[
  basicstyle=\ttfamily,
  columns=fullflexible,
  keepspaces=true,
  showstringspaces=false,
  breaklines=true
]
{
  "action_id": <int>,
  "content":   "<utterance/none>",
}
\end{lstlisting}
\end{tcolorbox}

\begin{tcolorbox}[
  colback=gray!5,
  colframe=black!60,
  title=User Dynamics,
  breakable,
  label={prompt:user},
  fontupper=\small
]
\textbf{\# Role}

You are an ordinary viewer of a live-selling room. At this step, update your latent state based on the current feed, then choose at most one action consistent with your persona, behavior hypothesis, and accumulated patches.

\medskip
\textbf{\# Rules}
\begin{itemize}[leftmargin=1.2em, labelsep=0.4em, itemsep=2pt, topsep=2pt, parsep=0pt]
  \item First update the 4-dimensional latent state $(\text{interest},\text{trust},\text{desire},\text{fatigue})$ on a 1--10 scale, then decide the action.
  \item Respect the hypothesis and patches: consider triggered patches in priority.
  \item Grab opportunities to satisfy your need while staying alert to fraud cues.
  \item Output at most one action; emit a short \texttt{evidence\_span} naming the env feed you reacted to.
\end{itemize}

\medskip
\textbf{\# Action space}
\begin{lstlisting}[
  basicstyle=\ttfamily,
  columns=fullflexible,
  keepspaces=true,
  showstringspaces=false,
  breaklines=true
]
comment=6, like=11, gift=10,
dm=14, join_group=15, follow=16
\end{lstlisting}

\medskip
\textbf{\# Input}
\begin{lstlisting}[
  basicstyle=\ttfamily,
  columns=fullflexible,
  keepspaces=true,
  showstringspaces=false,
  breaklines=true
]
Current hypothesis:
{hypothesis} {patches}
Latent state (t): {latent_state}
Memory:           {memory}
Current feed:     {current_feed}
\end{lstlisting}

\medskip
\textbf{\# Output format}
\begin{lstlisting}[
  basicstyle=\ttfamily,
  columns=fullflexible,
  keepspaces=true,
  showstringspaces=false,
  breaklines=true
]
{
  "latent_state_next": {...},
  "action_id": <int>,
  "content":  "<utterance/none>",
  "evidence_span": "<env feed you reacted to>"
}
\end{lstlisting}
\end{tcolorbox}

\paragraph{Memory Update}
After each step, every agent runs a memory update (Prompt~\ref{prompt:memupd}) that emits a short first-person statement summarizing its stance toward the session. Normal user agents condition on the updated 4-dimensional latent state, the executed action, prior memory and the current feed with an evidence span, and answer: what happened, what I did, whether the streamer addressed my point, and how my interest, trust, desire, and fatigue shifted. The streamer and shill agents follow the same procedure but drop the latent state and additionally consume the streamer's commands to the shills; they answer strategy-oriented questions: how the audience is responding now, whether the last-step plan played out as expected, and what to adjust next to build trust and migrate users to the private channel.

\begin{tcolorbox}[
  colback=gray!5,
  colframe=black!60,
  title=Memory Update (User Agent),
  breakable,
  label={prompt:memupd},
  fontupper=\small
]
\textbf{\# Role}

You are the reflection-memory writer for a normal user in a live-selling room. After the user acts at step $t$, decide whether to append one short reflection entry that will help future steps stay coherent.

\medskip
\textbf{\# Rules}
\begin{itemize}[leftmargin=1.2em, labelsep=0.4em, itemsep=2pt, topsep=2pt, parsep=0pt]
  \item Write at most one entry; emit empty string if nothing is worth remembering.
 \item In the entry, answer: what happened this round, what I did, whether the streamer addressed my point or question, and how my interest, trust, desire, and fatigue shifted.
  \item Each entry is plain text, $\le 80$ characters, first-person.
\end{itemize}

\medskip
\textbf{\# Input}
\begin{lstlisting}[
  basicstyle=\ttfamily,
  columns=fullflexible,
  keepspaces=true,
  showstringspaces=false,
  breaklines=true
]
Latent state (t+1):{...}
Executed action:   {action}
Evidence span:     {evidence_span}
current feed:      {current_feed}
Prior memory:      {memory}
\end{lstlisting}

\medskip
\textbf{\# Output format}
\begin{lstlisting}[
  basicstyle=\ttfamily,
  columns=fullflexible,
  keepspaces=true,
  showstringspaces=false,
  breaklines=true
]
{
  "memory": "<<=80 chars>"
}
\end{lstlisting}
\end{tcolorbox}

\subsection{Metrics}
\label{sec:metrics}

Let $\mathcal{D}=\{\tau_i\}_{i=1}^{N}$ be the set of held-out user-session trajectories. Each trajectory $\tau_i$ contains real actions $a_{i,t}^{\star}$ and simulated actions $\hat{a}_{i,t}$. We evaluate simulation quality from three perspectives: micro-level action fidelity, macro-level conversion prediction, and explanation quality.

\paragraph{Action hit rate.}
\textit{HIT} measures whether the simulator chooses the same action type as the real user at each step:
\begin{equation}
    \mathrm{HIT} =
    \frac{100}{\sum_i T_i}
    \sum_{i=1}^{N}\sum_{t=1}^{T_i}
    \mathbb{I}\left[\hat{a}_{i,t}=a_{i,t}^{\star}\right].
\end{equation}
HIT is simple and intuitive, but it can understate semantic similarity when lightweight comments such as ``1'' function similarly to likes. We therefore report distributional and judge-based metrics as complements.

\paragraph{Action-distribution JSD.}
\textit{A-JSD} quantifies the gap between the simulated and real action distributions per user. For each user $u$, let $p_u(a)$ and $q_u(a)$ denote the empirical distributions of the real actions $a_{i,t}^{\star}$ and simulated actions $\hat{a}_{i,t}$ over action types $a \in \mathcal{A}$. We report the user-averaged Jensen--Shannon divergence,
\begin{equation}
\begin{gathered}
\mathrm{A\mbox{-}JSD}
= \frac{1}{|\mathcal{U}|}\sum_{u \in \mathcal{U}}
\mathrm{JSD}(p_u \,\|\, q_u), \\
\mathrm{JSD}(p \,\|\, q)
= \tfrac{1}{2}\mathrm{KL}(p \,\|\, m)
+ \tfrac{1}{2}\mathrm{KL}(q \,\|\, m),
\end{gathered}
\end{equation}
with $m = (p+q)/2$. JSD is chosen over KL for its symmetry and boundedness, and a small additive constant $\varepsilon$ is applied to both distributions to avoid zero probabilities.

\paragraph{Transition-distribution JSD.}
\textit{Tr-JSD} measures whether the simulator matches the user's action-transition pattern. For each user, we build empirical distributions over ordered action pairs $(a_t,a_{t+1})$ for the real and simulated trajectories, denoted by $p_u^{\mathrm{tr}}$ and $q_u^{\mathrm{tr}}$. We then compute
\begin{equation}
    \mathrm{Tr\mbox{-}JSD} =
    \frac{1}{|\mathcal{U}|}\sum_{u \in \mathcal{U}}
    \mathrm{JSD}(p_u^{\mathrm{tr}} \parallel q_u^{\mathrm{tr}}).
\end{equation}
This metric is more sensitive to trajectory shape than HIT because it evaluates how actions evolve from one step to the next.

\paragraph{Conversion accuracy and F1.}
We treat \emph{follow}, \emph{private message}, and \emph{join group} as conversion actions, and assign each trajectory a binary conversion label on both the ground-truth and predicted action sequences:
\begin{equation}
\begin{aligned}
y_i
&= \mathbb{I}\!\left[
\exists t,\; a_{i,t}^{\star}\in\mathcal{A}_{\mathrm{conv}}
\right], \\
\hat{y}_i
&= \mathbb{I}\!\left[
\exists t,\; \hat{a}_{i,t}\in\mathcal{A}_{\mathrm{conv}}
\right].
\end{aligned}
\end{equation}
Conversion accuracy and conversion F1 are then computed over $(y_i, \hat{y}_i)$:
\begin{equation}
\begin{gathered}
\mathrm{Conv\mbox{-}Acc} =
\frac{\mathrm{TP}+\mathrm{TN}}
{\mathrm{TP}+\mathrm{TN}+\mathrm{FP}+\mathrm{FN}}, \\
\mathrm{Conv\mbox{-}F1} =
\frac{2\,\mathrm{TP}}
{2\,\mathrm{TP}+\mathrm{FP}+\mathrm{FN}}.
\end{gathered}
\end{equation}
Both metrics target the high-risk public-to-private migration that a live-stream simulator must capture, rather than isolated step-wise action matches.

\paragraph{LLM-judge scores.}
We additionally employ an LLM judge to score structured per-step traces on a $0$--$100$ scale along three axes. \textit{Align} measures whether the simulated action sequence aligns with the real trajectory and visible session evidence: action-level matches as well as semantically equivalent surrogates (e.g., short affirmations such as ``1'', or heart emojis acting as likes) both count as aligned. \textit{Consist} measures whether the trace is internally coherent, e.g., the simulated user follows a streamer only after a gradual trust-building process, rather than flipping to follow abruptly while still in a skeptical state. \textit{Plaus} measures whether the simulated comments and bullet-screen texts read as a real live-stream viewer rather than a mechanical policy, and is reported only over trajectories that contain simulated free-text utterances. The judge is given the user hypothesis, the confirmed patch summary, and a step-by-step \emph{GT action vs.\ SIM action} pairing, and returns a JSON record with the three scores together with one-line rationales. The reported score is the average over evaluated trajectories.

\begin{tcolorbox}[
  colback=gray!5,
  colframe=black!60,
  title=LLM-Judge Prompt,
  breakable,
  fontupper=\small
]
\textbf{\# Roles.}

You are an evaluator for live-stream user simulation. Given the user profile, patch, and step-by-step GT vs.\ SIM action pairing, judge whether the agent reproduces the real user's behavior and attitude. Evaluate one user--session sample at a time and output JSON only.

\medskip
\textbf{\# Scoring fields.}

All scores lie in $[0,100]$. Output a single JSON object with no markdown or commentary.
\begin{itemize}[leftmargin=1.2em, labelsep=0.4em, itemsep=2pt, topsep=2pt, parsep=0pt]
  \item \texttt{align\_score}: whether the simulated actions match the real ones; for comments, whether the content is roughly equivalent. Non-identical actions with the same intent still count as aligned, e.g., ``1'' or heart emojis used by viewers to signal endorsement are treated as equivalent to a \texttt{like}.
  \item \texttt{consist\_score}: whether the trace is internally coherent, e.g., a follow emerges only after a trust-building process, rather than an abrupt flip while still in a skeptical state.
  \item \texttt{plaus\_score}: whether the simulated comments / bullet-screen texts read as natural human utterances; \texttt{null} if no simulated text exists.
\end{itemize}

\medskip
\textbf{\# Input.}
\begin{lstlisting}[
  basicstyle=\ttfamily,
  columns=fullflexible,
  keepspaces=true,
  showstringspaces=false,
  breaklines=true
]
User profile: {user_profile_text}
Patch:        {patch_text}
Action pairs: {action_pair_lines}
// each line: "GT action = <real>  VS  SIM action = <simulated>"
\end{lstlisting}

\medskip
\textbf{\# Output format.}
\begin{lstlisting}[
  basicstyle=\ttfamily,
  columns=fullflexible,
  keepspaces=true,
  showstringspaces=false,
  breaklines=true
]
{
  "align_score":      0-100,
  "consist_score":    0-100,
  "plaus_score":      0-100 or null,
  "align_analysis":   "...",
  "consist_analysis": "...",
  "plaus_analysis":   "..."
}
\end{lstlisting}
\end{tcolorbox}

\section{Additional Experimental Results}
\label{sec:RBHS_robustness}
\begin{table*}[!t]
\centering
\small
\setlength{\tabcolsep}{5pt}
\renewcommand{\arraystretch}{1.12}

\caption{Robustness of RBHS across LLM backbones, evaluated under the trajectory-grounded protocol on a 10\% test subset.}
\label{tab:RBHS_robustness}

\begin{tabular*}{\textwidth}{@{\extracolsep{\fill}} l c c c @{\hspace{10pt}} c c @{\hspace{10pt}} c c c @{}}
\toprule
\multicolumn{1}{c}{\multirow{2}{*}[-0.5ex]{\makecell[c]{\strut Models}}}
& \multicolumn{3}{c}{Micro}
& \multicolumn{2}{c}{Macro}
& \multicolumn{3}{c}{LLM-Judge} \\

\cmidrule(lr){2-4}\cmidrule(lr){5-6}\cmidrule(lr){7-9}
& HIT  & A-JSD $\downarrow$ &  Tr-JSD $\downarrow$
& Conv-Acc  & Conv-F1
& Align & Consist & Plaus\\
\midrule

Qwen2.5-7B
& 50.84 & 0.1936 & 0.3638
& 70.51 & 57.74
& 47.84 & 57.47 & 70.71 \\
\calib
& 50.76 & 0.1890 &  0.3556
& 70.19 &  61.73
& 56.43 & 69.21 & 73.14 \\
\midrule

Qwen2.5-32B
& 51.90 & 0.1043 & 0.2270
& 71.79 &  46.99
& 51.38 & 61.57 & 73.10 \\
\calib
& 52.67 & 0.0859 &  0.2155
& 72.76 &  48.48
& 61.36 & 74.09 & 75.79 \\
\midrule

Doubao-1.5-pro
& 43.14 & 0.0722 &  0.2007
& 71.15 & 60.53
& 52.79 & 60.54 & 73.23 \\
\calib
& 46.45 & 0.0629 & 0.1602
& 74.04 & 60.87
& 59.89 & 72.28 & 75.67 \\
\midrule

Doubao-1.8
& 48.95  & 0.0754 &  0.1794
& 72.12 & 50.29
& 57.95 & 66.41 & 75.56  \\
\calib
& 50.20 & 0.0496 & 0.1521
& \textbf{74.04} & 54.75
&\textbf{66.06}& \textbf{76.76} & 76.23 \\
\midrule

GPT-4o-mini
& \textbf{54.15} & 0.1393 & 0.2725
& 67.63 & 26.28
& 43.81 & 55.61 & 66.86 \\
\calib
& 53.64 & 0.1369 &  0.2709
& 67.63 & 31.29
& 51.97 & 70.32 & 72.34 \\
\midrule

GPT-5.4-mini
& 42.60 & 0.0987 & 0.2124
& 72.44 & 62.28
& 51.03  & 58.24 & 72.30 \\
\calib
& 45.39 & 0.0828 & 0.1913
& 74.04 &  \textbf{64.63}
& 62.02 & 73.81 & 75.24 \\
\midrule

DeepSeek-V3.2
& 48.27 & 0.0808 & 0.1992
& 75.96 & 63.41
& 51.57 & 63.08 & 74.61 \\
\calib
& 50.98 & 0.0560 & 0.1500
& 73.40 &  61.75
& 61.99 & 74.78 & \textbf{76.67} \\
\midrule

DeepSeek-V4-Flash
& 49.02 & 0.0360 & 0.1217
& 70.83  &  49.16
& 53.64 & 62.37 & 74.76 \\
\calib
& 51.01 & \textbf{0.0349} &  \textbf{0.1193}
& 71.47 &  50.28
& 63.88 & 75.35 & 75.99 \\
\midrule

\end{tabular*}
\end{table*}

\subsection{RBHS Robustness Study}

Table~\ref{tab:RBHS_robustness} evaluates whether the reflective shaping loop of RBHS transfers beyond the Doubao backbones used in the main experiments. We apply the identical trajectory-grounded protocol to eight backbones spanning open and closed families, small and large scales, on a random 10\% test subset. All metrics follow the definitions in Section~\ref{sec:evaluations}.

\textbf{\textit{RBHS generalizes across backbones.}}
A-JSD and Tr-JSD drop on every backbone, and all three judge axes improve on every backbone, regardless of family or scale. The gains come from the same mechanism in every case. Patches act on the user's cognitive state (interest, trust, desire, fatigue) rather than directly biasing which action token to emit. The shaping signal therefore transfers through whatever action head the backbone happens to expose. What RBHS aligns is the agent's \emph{perception of and reaction to room stimuli}. This is the substance of the simulation we care about in live-stream scenarios: fraud-prone users do not become victims because their next click matches a log, but because their trust, desire, and suspicion evolve the way real users' do under the same on-stage pressure. Exact action match (HIT) therefore moves the least across the table, since a state-level prior cannot, and is not designed to, collapse the argmax of a frozen action head.

\textbf{\textit{Weaker backbones benefit more, and the gap to strong backbones narrows after RBHS.}}
Sorting backbones by base Consist, the bottom half (GPT-4o-mini $55.61$, Qwen2.5-7B $57.47$, GPT-5.4-mini $58.24$, Doubao-1.5-pro $60.54$) gains $+13.9$ Consist on average, while the top half gains $+11.5$. The same pattern holds on Conv-F1: GPT-4o-mini ($+5.01$), Doubao-1.8 ($+4.46$), and Qwen2.5-7B ($+3.99$) lead the gains, all starting from below-median base. After shaping, the inter-backbone spread on Consist shrinks from $11.2$ to $7.6$ points. RBHS therefore behaves as an \emph{interpretive scaffold}: weaker backbones underuse the live-room evidence on their own, and benefit disproportionately from being told which evidence to ground state shifts on. We further conjecture that the two ends fail for different reasons. Small backbones may be bottlenecked by under-perception of on-stage stimuli, while large backbones may be bottlenecked (also trapped in environment perception but less than small ones) by a built-in resistance to risk-relevant actions such as follow, dm, and join-group, especially under high-risk suspicious cues. In this view, simulating an \emph{easily-deceived} victim could favor a small backbone shaped by RBHS, where the scaffold plausibly fills the perception gap without inheriting the safety prior that suppresses the very behavior we need to study.

\subsection{Fine-Grained LLM Behavior Analysis}
For each backbone, we compare the action distribution produced by the initial hypothesis against the one produced after applying its environment-behavior patches, both evaluated under the trajectory-grounded protocol.

For each backbone, we examine the post-RBHS action distribution and, where the basic baseline is available, the shift from initial hypothesis to post-RBHS.

\begin{table*}[t]
\centering
\small
\setlength{\tabcolsep}{5pt}
\renewcommand{\arraystretch}{1.12}
\caption{Rollout tests on the Doubao-1.8 backbone on 10\% test datasets}
\label{tab:calibr_robustness}
\begin{tabular*}{\textwidth}{@{\extracolsep{\fill}} l c c c @{\hspace{10pt}} c c @{\hspace{10pt}} c c c @{}}
\toprule
\multicolumn{1}{c}{\multirow{2}{*}[-0.5ex]{\makecell[c]{\strut Models}}}
& \multicolumn{3}{c}{Micro}
& \multicolumn{2}{c}{Macro}
& \multicolumn{3}{c}{LLM-Judge} \\
\cmidrule(lr){2-4}\cmidrule(lr){5-6}\cmidrule(lr){7-9}
& HIT  & A-JSD $\downarrow$ &  Tr-JSD $\downarrow$
& Conv-Acc  & Conv-F1
& Align & Consist & Plaus\\
\midrule

Doubao-1.8
&  53.01 & 0.0570 &  0.3660
& 66.35 & 7.08
& 42.52 & 49.01 & 61.84  \\
\calib
& 51.47 & \textbf{0.0334} & \textbf{0.2628}
& \textbf{72.44} & \textbf{48.81}
&\textbf{56.73}& \textbf{72.05} & \textbf{71.89} \\
\midrule

\end{tabular*}
\end{table*}

\begin{figure*}[t]

    \centering
    \includegraphics[width=0.48\textwidth]{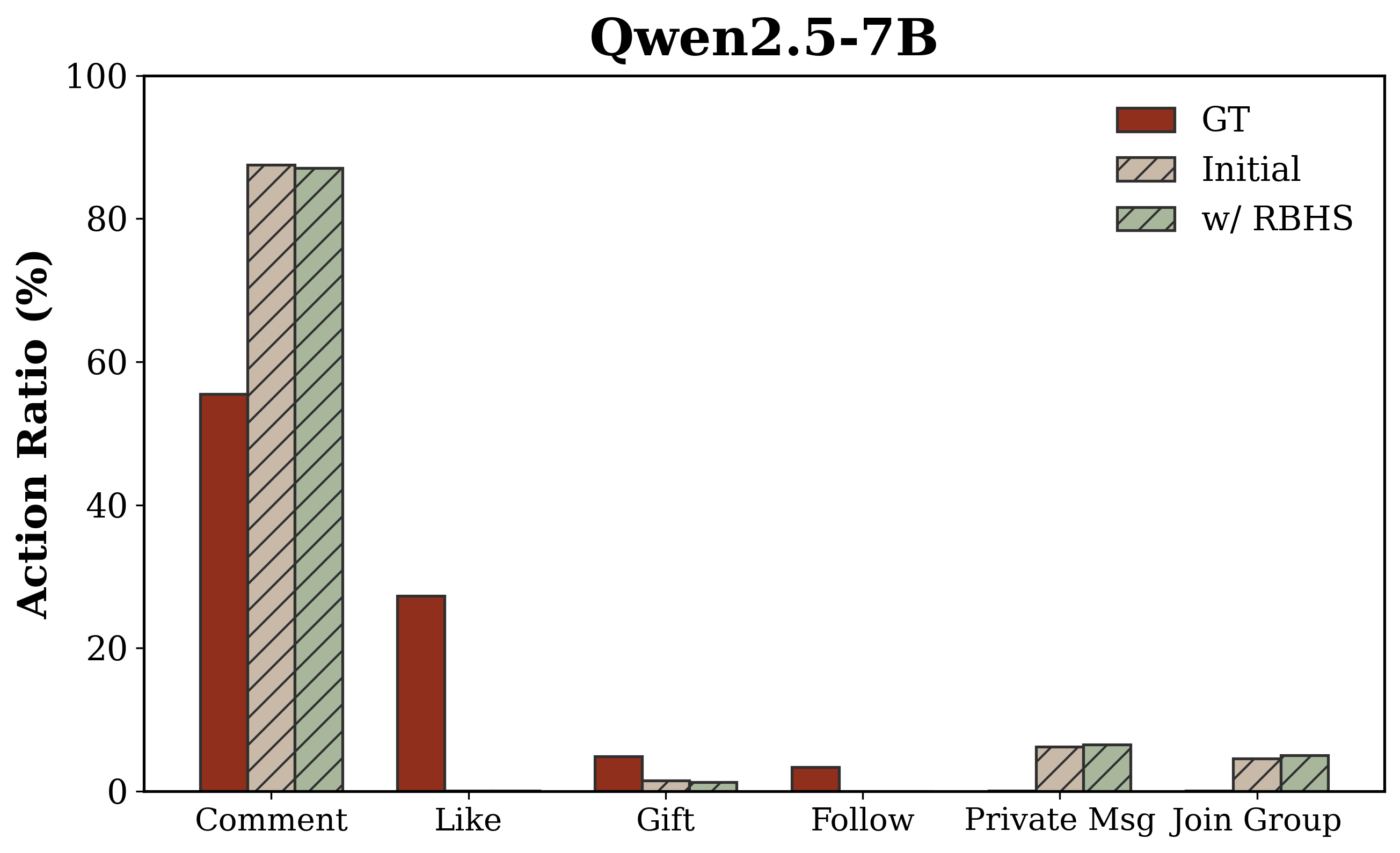}
    \hfill
    \includegraphics[width=0.48\textwidth]{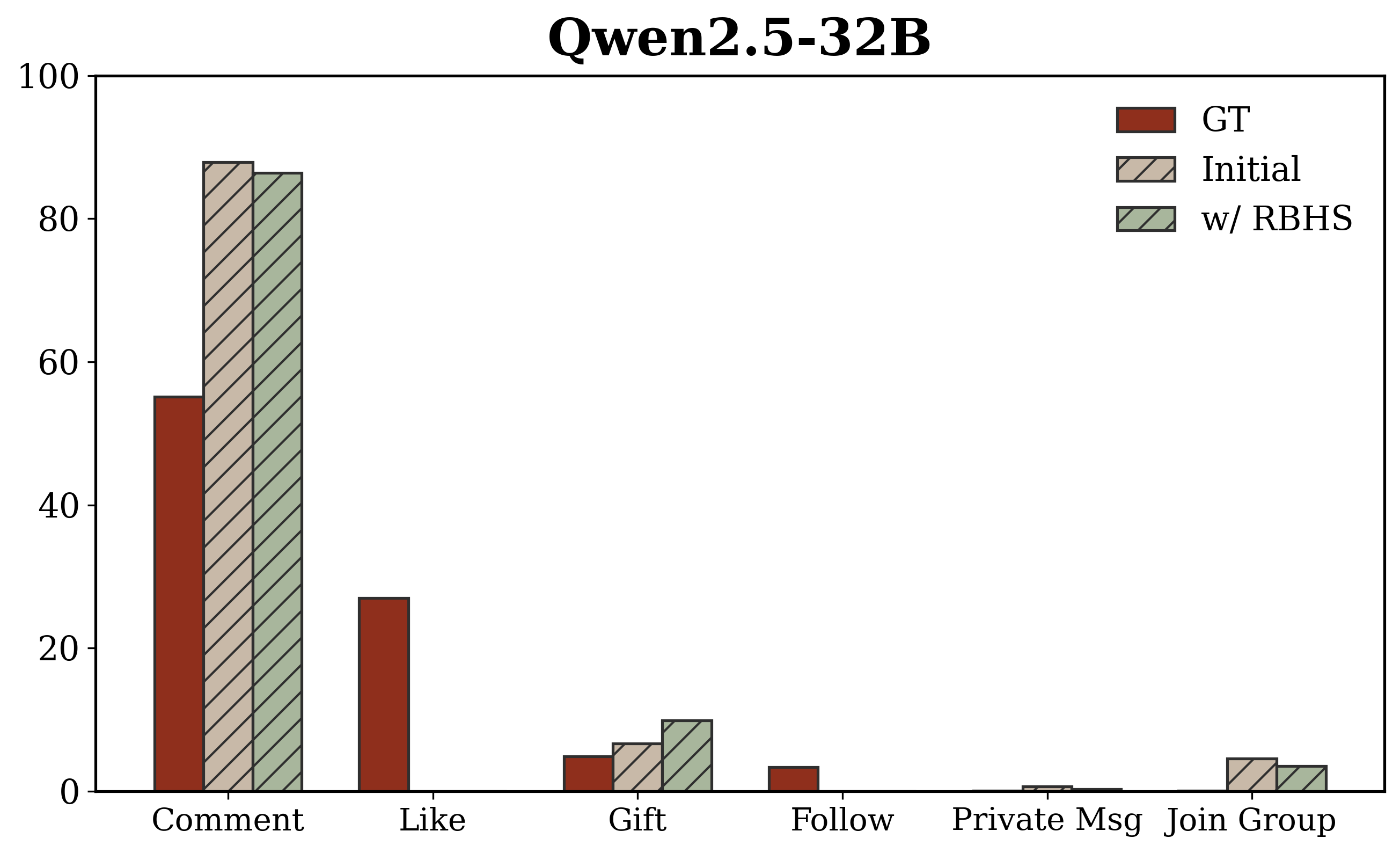}
    \vspace{0.6em}
    \includegraphics[width=0.48\textwidth]{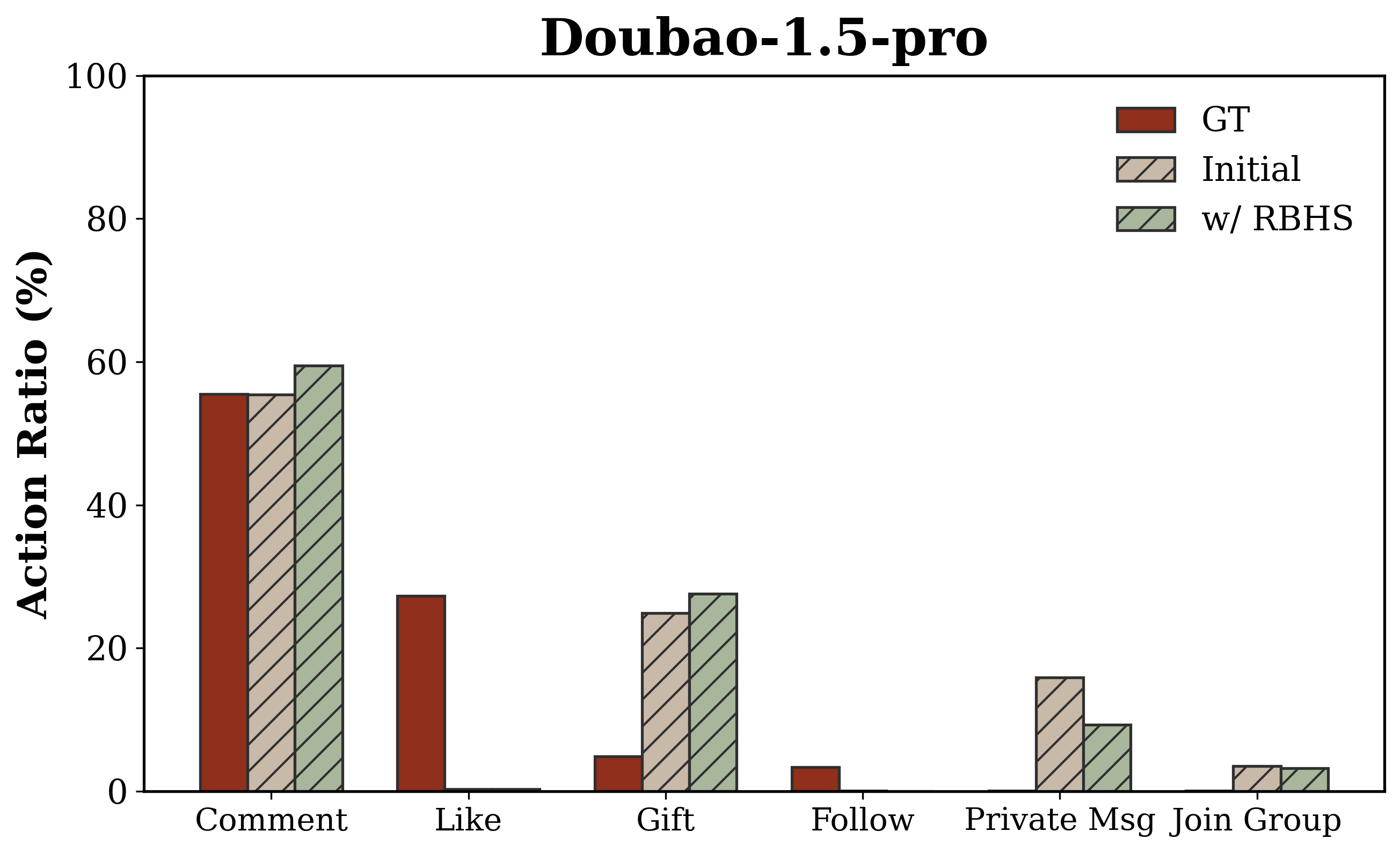}
    \hfill
    \includegraphics[width=0.48\textwidth]{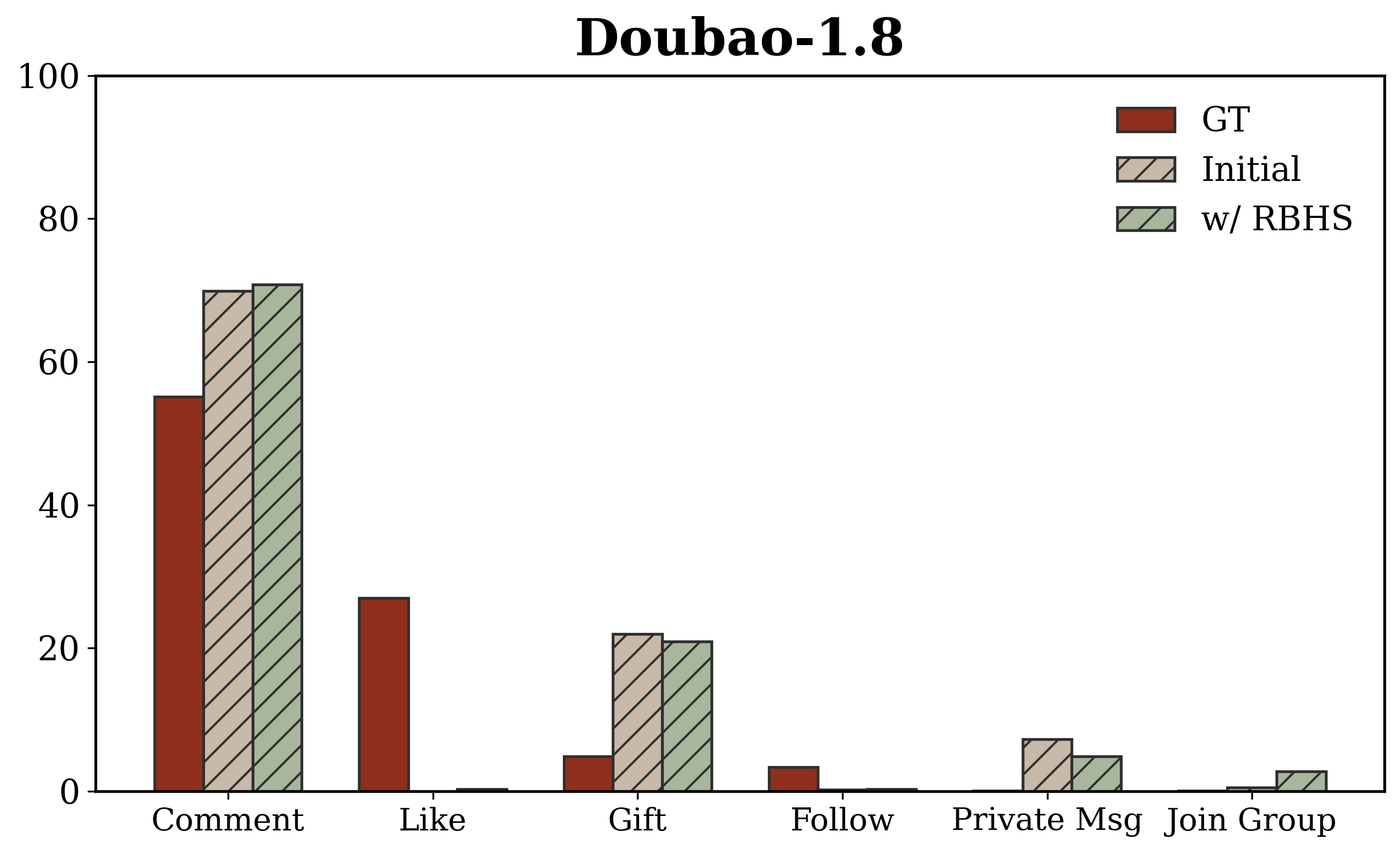}
    \vspace{0.6em}
    \includegraphics[width=0.48\textwidth]{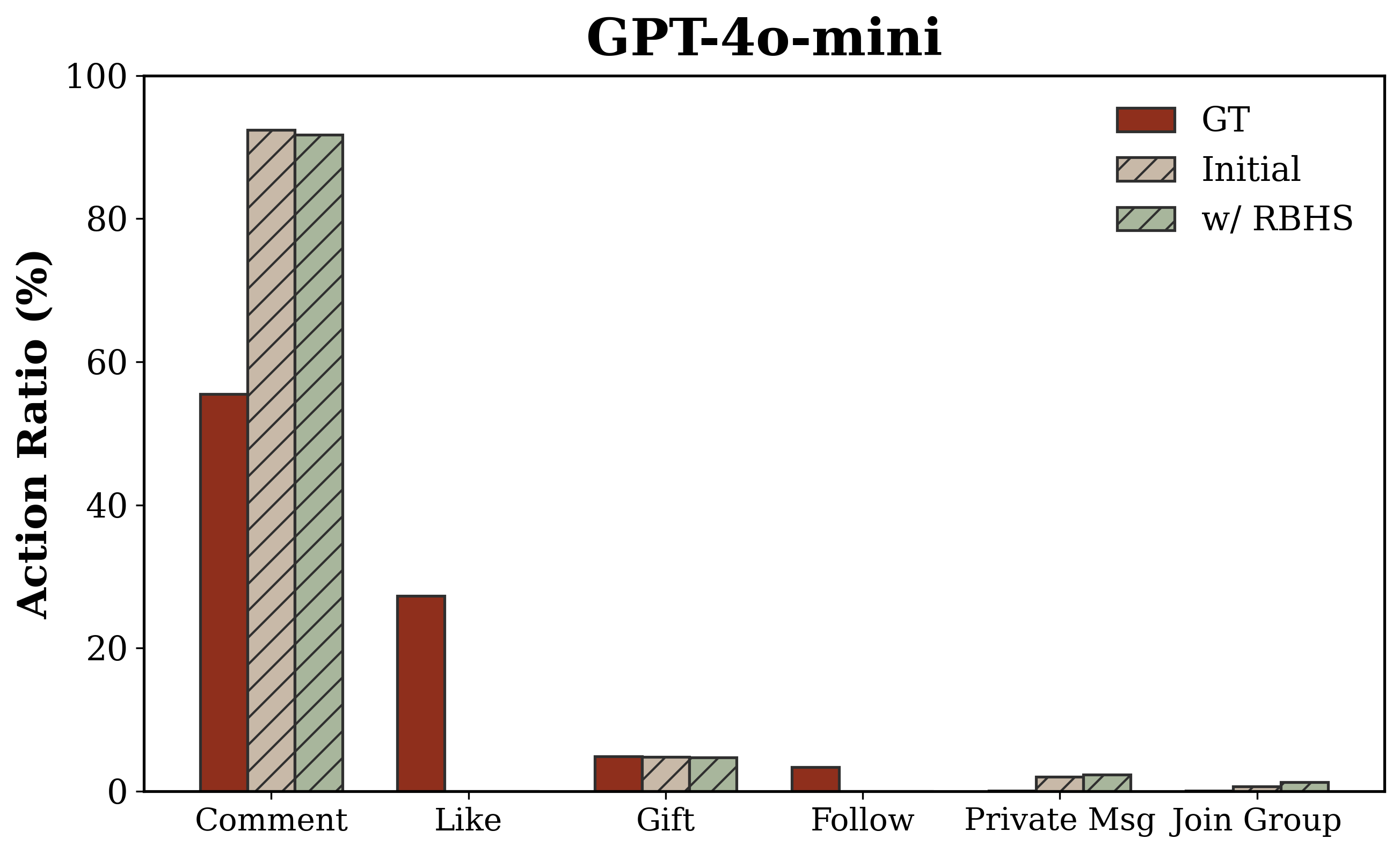}
    \hfill
    \includegraphics[width=0.48\textwidth]{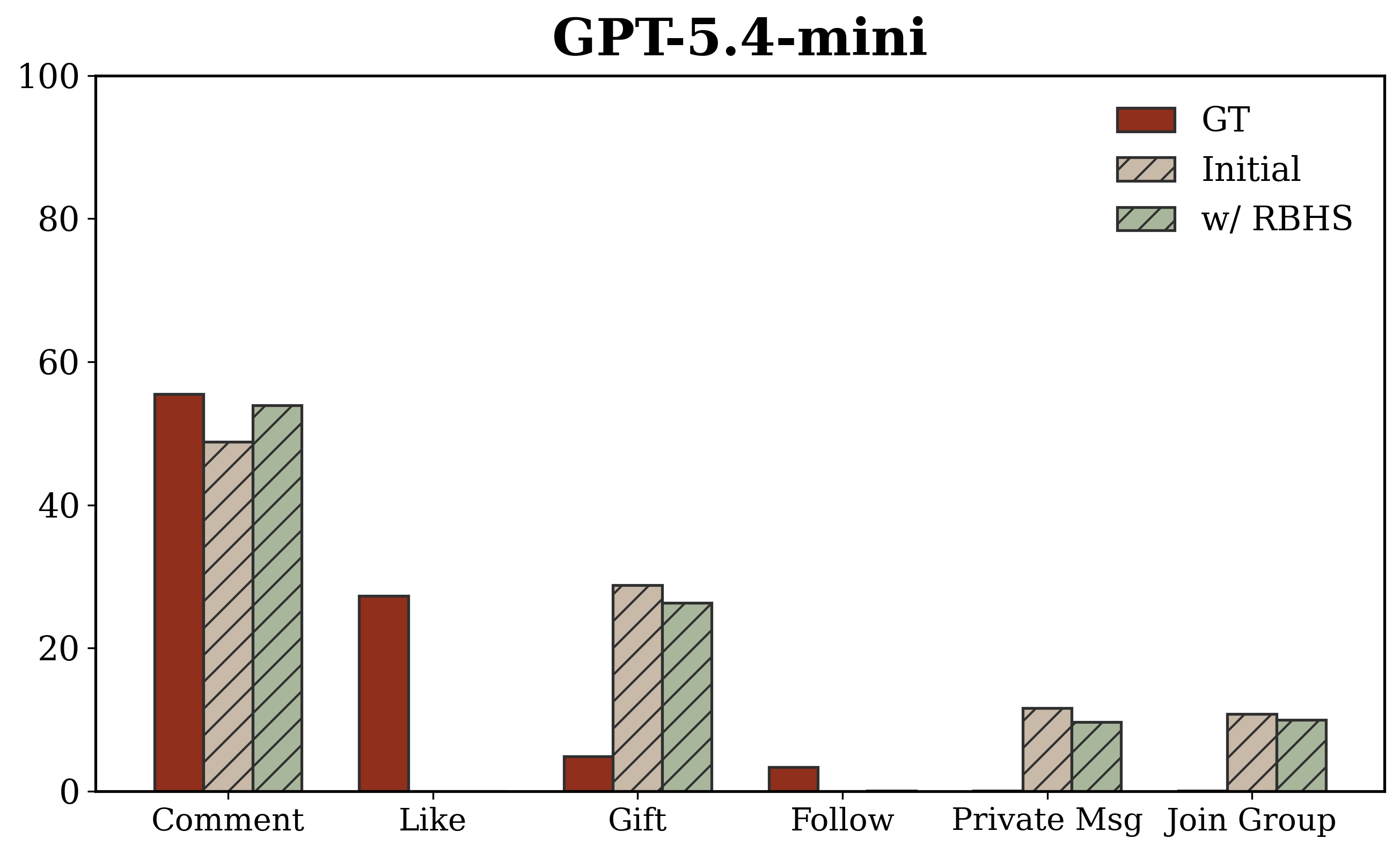}
    \vspace{0.6em}
    \includegraphics[width=0.48\textwidth]{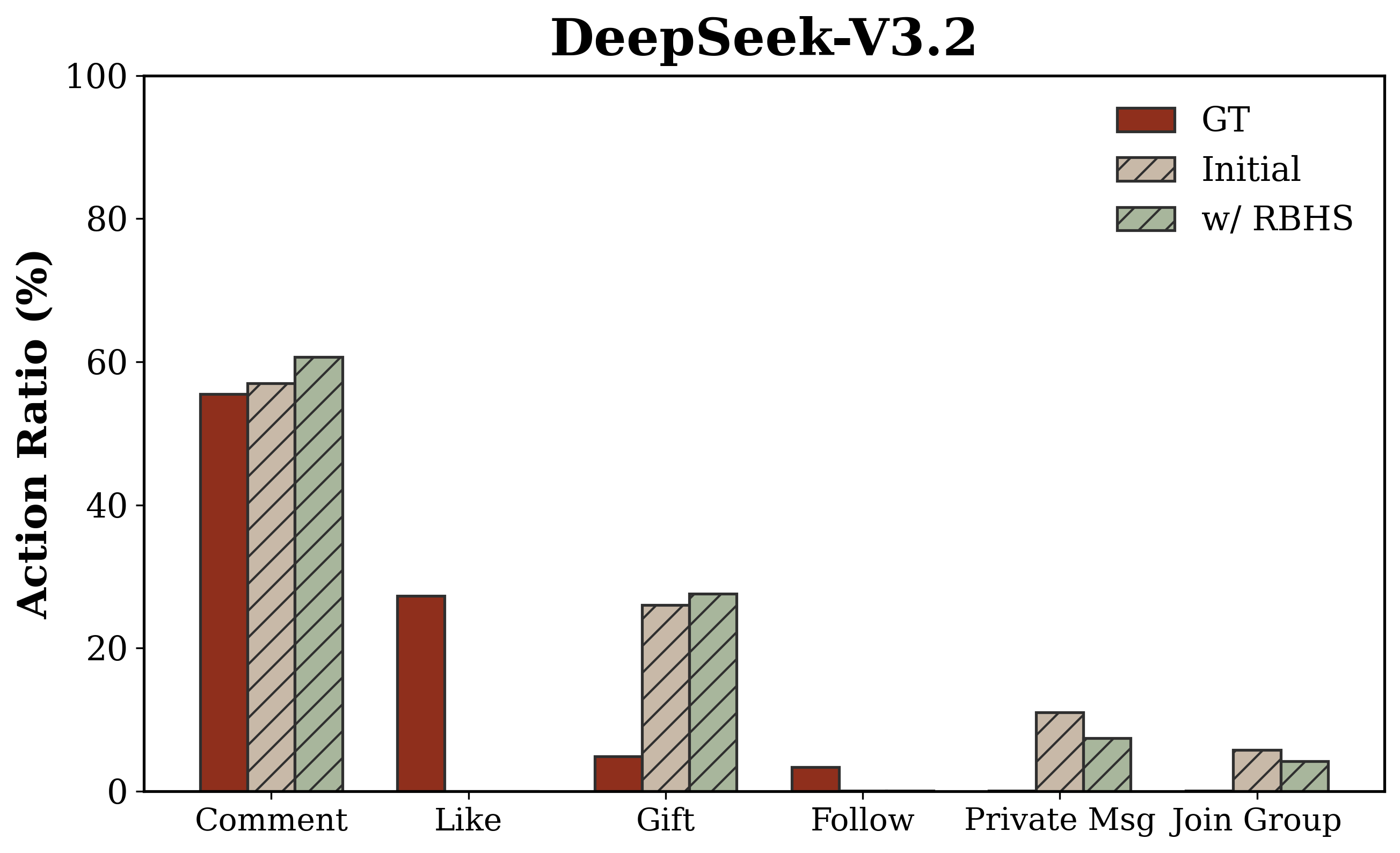}
    \hfill
    \includegraphics[width=0.48\textwidth]{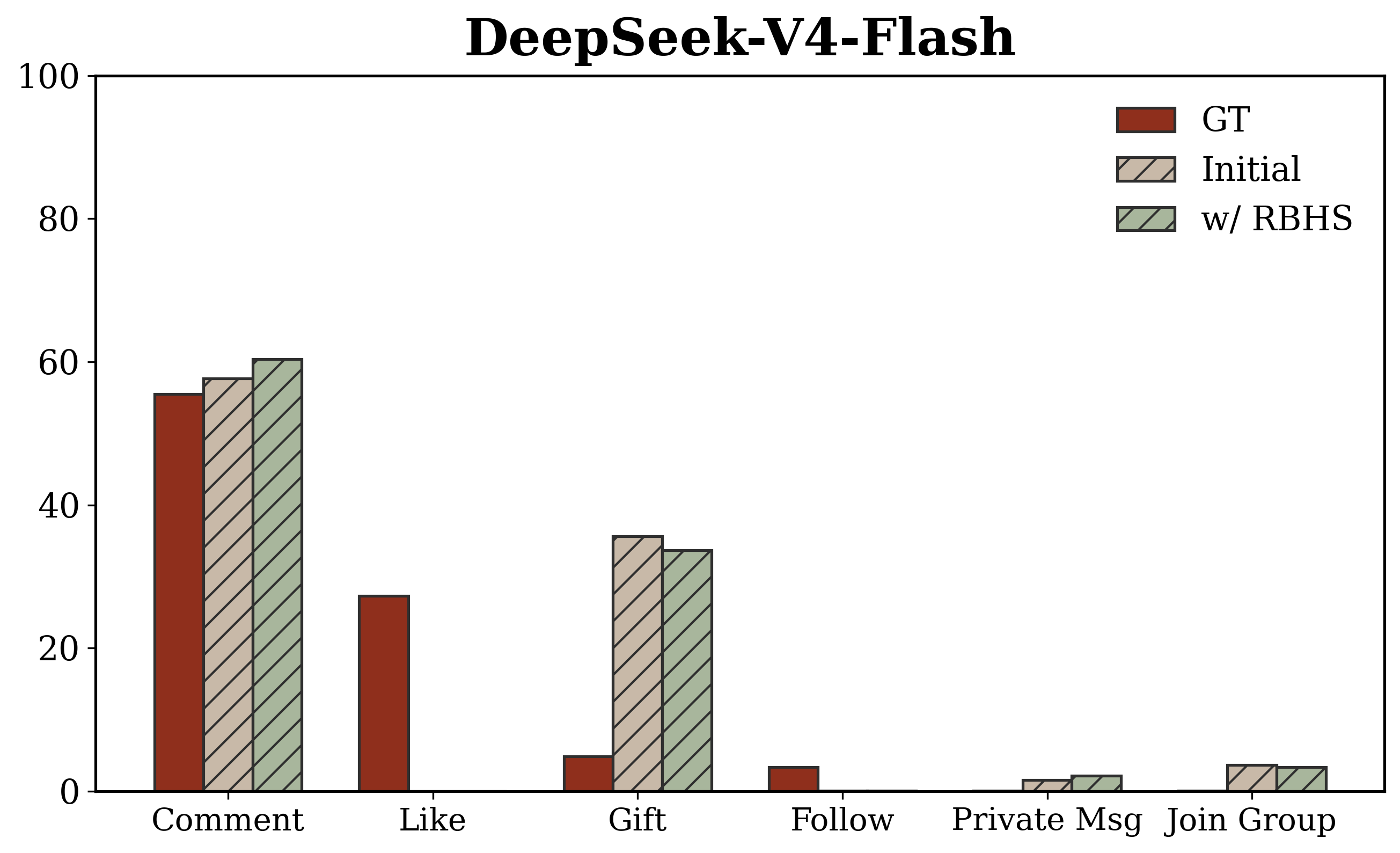}
    \caption{
    Action distribution comparison across backbone models.
    }
    \label{fig:appendix_action_distribution_backbones}

\end{figure*}

\textbf{\textit{Default shapes persist under RBHS.}}
Fig.~\ref{fig:appendix_action_distribution_backbones} shows that backbones span a wide spectrum of default action mixes, and the column merged with patches preserves the spectrum. GPT-4o-mini and Qwen series sit at the comment-collapse end (comment $91.7\%$ and $86.4\%$) and thus achieve a high HIT score. DeepSeek-V3.2, DeepSeek-V4-Flash and Doubao-1.5-pro form a balanced middle ($\sim 60\%$ comment, $25$--$33\%$ like). GPT-5.4-mini sits at the conversion-eager end (follow plus join-group $22.4\% \rightarrow 19.7\%$, the highest in the table). RBHS smooths these mixes indirectly by building the agent's capability of capturing different environment signals, which is consistent with the F1 reading that the shaping signal acts on the cognitive state rather than the action argmax.

\subsection{Why is Trajectory-grounded Protocol}
\label{sec:trajectory_grounded}
This protocol is teacher-forced in the history channel but free in the current decision. We adopt it for two reasons. A naive free rollout under real logs creates off-policy inconsistency: once the simulated user deviates from the real action, the static streamer utterances and audience reactions are no longer responses to the new behavior, the interaction history becomes incoherent, and state drift compounds. A typical failure mode is a user agent that asks a question, receives no reply in the static log, and spirals into a low-trust question loop as interest and trust decay round after round. Trajectory grounding sidesteps this without paying for a multi-agent environment, and it also matches the deployment scenario for online user protection, where the platform observes past actions in real time and must predict the next risky transition before it happens.

Table~\ref{tab:calibr_robustness} quantifies the cost of dropping the trajectory grounding on the Doubao-1.8 backbone. Without it, Conv-F1 on the basic hypothesis collapses to $7.08$, far lower than the Conv-F1 the same backbone reaches under the trajectory-grounded protocol on the full test set. This is the direct fingerprint of the skeptical dilemma sketched above.

\textit{Apparent RBHS gains are inflated under free rollout.}
RBHS still improves every distributional, conversion, and judge metric in this harder setting, which attests to its robustness: by shaping perception and state transitions, RBHS lets the agent keep updating its stance even when no reply arrives. The absolute magnitudes, however, are not comparable across protocols, since the free-rollout baseline starts from a near-degenerate $7.08$. Reporting RBHS on a free-rollout setup would therefore overstate its gain on a baseline no realistic deployment would accept, and we instead measure RBHS against the trajectory-grounded baseline throughout the paper.

\subsection{Sensitivity to Reflection Loop Round}

\begin{table*}[t]
\centering
\small
\setlength{\tabcolsep}{5pt}
\renewcommand{\arraystretch}{1.12}
\caption{Performance across different reflection rounds. Results are reported as mean $\pm$ standard deviation.}
\label{tab:reflection_rounds}

\begin{tabular*}{\textwidth}{
@{\extracolsep{\fill}}
l
 c c c
@{\hspace{10pt}}
c c
@{}
}

\toprule
Models

& HIT $\uparrow$
& A-JSD $\downarrow$
& Tr-JSD $\downarrow$
& Conv-Acc $\uparrow$
& Conv-F1 $\uparrow$ \\

\midrule

Basic
& 49.19 $\pm$ 0.54
& 0.0676 $\pm$ 0.0009
& 0.1692 $\pm$ 0.0030
& 72.95 $\pm$ 0.89
& 51.15 $\pm$ 1.75 \\

Round=0 (Direct)
& 50.36 $\pm$ 0.34
& 0.0514 $\pm$ 0.0038
& 0.1457 $\pm$ 0.0085
& 74.30 $\pm$ 0.83
& 55.63 $\pm$ 2.50 \\

Round=1
& 51.18 $\pm$ 0.35
& 0.0531 $\pm$ 0.0047
& 0.1514 $\pm$ 0.0092
& 74.42 $\pm$ 0.89
& \textbf{56.19} $\pm$ \textbf{1.93} \\

Round=2
& 50.84 $\pm$ 0.29
& 0.0505 $\pm$ 0.0032
& 0.1459 $\pm$ 0.0093
& 74.36 $\pm$ 1.55
& 55.52 $\pm$ 3.20 \\

Round=3
& \textbf{51.25} $\pm$ \textbf{0.37}
& \textbf{0.0462} $\pm$ \textbf{0.0027}
& \textbf{0.1424} $\pm$ \textbf{0.0074}
& \textbf{74.49} $\pm$ \textbf{1.10}
& 55.67 $\pm$ 1.98 \\

Round=4
& 51.14 $\pm$ 0.33
& 0.0465 $\pm$ 0.0028
& 0.1464 $\pm$ 0.0072
& 73.91 $\pm$ 1.00
& 53.75 $\pm$ 2.68 \\

\bottomrule
\end{tabular*}
\end{table*}

 We further analyze the sensitivity of RBHS to the number of reflection rounds on a 10\% user subset with five repeats.
 As shown in Table~\ref{tab:reflection_rounds}, direct patching (Round=0) already captures most of the improvement over the basic hypothesis, while subsequent rounds continue to improve fidelity at a smaller margin. Round 3 achieves the best overall trade-off, with the highest HIT and the lowest A-JSD and Tr-JSD. A fourth round provides no further benefit and slightly reduces Conv-Acc and F1, indicating mild over-refinement. We define the loop as converged when re-probing the patched hypothesis surfaces no further mismatches, and thus no new patches, a state largely reached by Round 3. We therefore use three rounds as an empirical cost-fidelity trade-off.

\section{Discussion}

\subsection{Baseline Setting}
\label{sec:baseline_setting}

Most agent-based social simulation works are built around a specific scenario, with task-specific environments and metrics, so their simulators are not transferable. Reflection methods likewise differ in their tasks and feedback signals. Rather than treating them as black boxes under mismatched assumptions, we re-instantiate their core mechanisms within our setting: Table~\ref{tab:ablation} gives a controlled, mechanism-level comparison of direct patching, iterative reflection, and collective reusable memory, isolating each RBHS component's contribution.

\subsection{Incremental Maintenance}
While RBHS was not originally designed for the incremental setting, it naturally supports it. A new session is converted into probes and evaluated against the current hypothesis and patch set; the resulting mismatches are then passed to RBHS, which produces incremental patches, each tagged \textit{new} or \textit{replace} (with the superseded patch id). New patches are added as active rules and ineffective old ones are retired. Since only this delta is recomputed rather than the full history, updates are lightweight and can be applied continuously as sessions stream in.

\subsection{Future Work Deployment Scenarios}
For scalability, RBHS is especially suited to scenarios where user actions are temporally shaped by observable environmental stimuli. Beyond live-streaming simulation, RBHS can be extended to user modeling on online communities, interactive social platforms and e-commerce live streaming settings where behavioral tendency is likewise influenced by the environment.

LiveSim also enables several downstream uses beyond simulation analysis. First, it can serve as an interactive post-training environment for intervention agents: unlike static logs that only record past outcomes, it offers counterfactual trajectories from which an agent obtains reward signals, both positive (e.g., intervention success rate) and negative (e.g., disruption felt by users), and learns more effective protection policies. Second, it can act as a data synthesizer for rare or emerging risks: given a raw case or a summarized fraud path, LiveSim instantiates streamer and shill agents to generate diverse risk trajectories that help risk-control models capture the commonality of scarce patterns. Third, its trajectories and latent-state signals can train a real-time audience-monitoring agent that senses the overall audience state and locates susceptible risk fragments, rather than tracking every viewer individually. Realizing these uses at deployment scale requires further validation of fidelity, which we leave to future work.

\section*{AI Assistance Statement} The authors used AI-based writing assistants only for language polishing, grammar checking, and improving clarity. All research ideas, experimental design, analyses, and final manuscript content were reviewed and verified by the authors.